\documentclass[extra,mreferee]{gji}

\usepackage{color}
\usepackage{timet}
\usepackage{amsmath}
\usepackage[pdftex]{graphicx}
\usepackage{subfig}

\usepackage{amsmath} 
\usepackage{amssymb,stmaryrd} 
\usepackage{amsfonts}
\usepackage{amscd}   

\usepackage{algorithm}
\usepackage{algorithmic}
\usepackage{setspace} 

\usepackage{gensymb}

\usepackage{slashbox}
\usepackage{pict2e}
\usepackage{multirow}

\def\RB{{\mathbb R}}
\def\FM{{\mathcal F}}
\def\JM{{\mathcal J}}
\def\OM{{\mathcal O}}
\def\SM{{\mathcal S}}
\newenvironment{definition}[1][Definition]{\begin{trivlist}
\item[\hskip \labelsep {\bfseries #1}]}{\end{trivlist}}

\begin{document}
\DeclareGraphicsExtensions{.pdf,.jpg,.mps,.png}

\title[Randomized Geologic Feature Detection]{Efficient Data-Driven Geologic 
Feature Detection from Pre-stack Seismic Measurements using \\Randomized Machine-Learning Algorithm}
\author[Lin et al.]
{\Large{Youzuo Lin}$^{\mathbf{1}, \mathbf{*}}$, Shusen Wang$^\mathbf{2}$,
Jayaraman Thiagarajan$^\mathbf{3}$, George Guthrie$^\mathbf{1}$, and David Coblentz$^\mathbf{1}$ \\
$\mathbf{1}:$ Earth and Environment Division, Los Alamos National Laboratory, Los Alamos, NM 87545;\\
$\mathbf{2}:$ Department of Statistics, University of California, Berkeley, CA 94720;\\
$\mathbf{3}:$ Center for Applied Scientific Computing, Lawrence Livermore National Laboratory, Livermore, CA 94550.\\
Correspondence $\mathbf{*}$: ylin@lanl.gov.}

\date{}
\volume{}
\pubyear{}

\label{firstpage}

\maketitle



\begin{summary}
Conventional seismic techniques for detecting the subsurface geologic features are challenged 
by limited data coverage, computational inefficiency, and subjective human factors. We developed 
a novel data-driven geological feature detection approach based on pre-stack  seismic measurements. 
Our detection method employs an efficient and accurate machine-learning detection approach to extract useful 
subsurface geologic features automatically. Specifically, our method is based on kernel ridge regression
model. The conventional kernel ridge regression can be computationally prohibited because of the large volume of seismic
measurements. We employ a data reduction technique in combination 
with the conventional kernel ridge regression method to improve the computational efficiency 
and reduce memory usage. In particular, we utilize a randomized numerical linear algebra 
technique, named Nystr\"om method,  to effectively reduce the dimensionality of the feature space without compromising the information 
content required for accurate detection. We provide thorough computational cost analysis to show efficiency 
of our new geological feature detection methods. We further validate the performance of our new subsurface geologic
feature detection method using synthetic surface seismic data for 2D acoustic and elastic velocity models. Our 
numerical examples demonstrate that our new detection method significantly improves the computational 
efficiency while maintaining comparable accuracy. Interestingly, we show that our 
method yields a speed-up ratio on the order of $\sim10^2$ to $\sim 10^3$ in a multi-core 
computational environment.

\end{summary}


\section{Introduction}

It is challenging to analyze and interpret seismic measurements for identifying prospective geological features.
The challenges arise from processing of large volumes of seismic data and subjective human factors. 
Different geologic features play different roles in characterizing the subsurface structure. Since geologic fault 
is one of the most interesting features in subsurface characterization, we use that as the target to demonstrate the efficacy 
of our new machine-learning based geologic feature detection method. Geologic fault zone is essential to 
various subsurface energy applications. In geothermal exploration, geologic faults provide important information 
for siting the drilling wells. In carbon sequestration, geologic faults can be critical to monitor the potential 
leaks of stored CO$_2$. In oil \& gas production, geologic faults are used to signal reservoir boundaries or 
hydrocarbon traps. 

In current seismic exploration, most subsurface characterization techniques are physics dominated, meaning that 
the governing physics equations are well understood and utilized to describe the underlying physics of the problems 
of interest. A well known examples of this is the seismic full-waveform inversion (FWI)~\citep{Acoustic-2015-Lin, 
Quantifying-2015-Lin, Virieux-2009-Overview}. In FWI, an inverse problem is formulated to connect the measurements 
and the governing physics equations. Numerical optimization techniques are utilized to solve for the subsurface models. 
Similar framework and procedures can be applied to many other techniques such as seismic imaging~\citep{Stable-2015-Zhang, Least-2015-Lin}, 
tomography~\citep{Lin-2015-Double, Seismic-2014-Rawlinson}, etc. With the numerical model built up from the optimization, human 
intervention is involved to further interpret and finalize the acceptable models. Even though those conventional methods 
have been shown great success in many applications, in some situations they can be limited because of poor data coverage, 
computational inefficiency, and subjective human factors. A robust, efficient, and accurate subsurface characterization 
method is therefore needed to address those needs.

Given the advances in data science and machine learning technologies, there has been a recent surge in utilizing automated 
data-driven algorithms to detect subsurface geologic 
features~\citep{Schnetzler-2017-Use, Araya-2017-Automated, Guillen-2015-Supervised, Zhang-2014-Machine, Methods-2013-Hale,
Ramirez-2011-Machine}. In seismic applications, those methods can be categorized into either learning-from-data or 
learning-from-model. The major differentiation between these two types of methods are whether the learning algorithms are employed 
on the pre-stack or post-stack seismic data sets. Most of the current existing work of machine learning methods for seismic applications are
based on the post-stack seismic data sets, meaning migrated or inverted models need to be obtain prior to the usage of machien learning
techniques. In~\cite{Guillen-2015-Supervised}, migration imaging models are first obtained from seismic data sets. Supervised 
learning technique is then applied to the imaging model to automatically detect the salt body. Similarly, in~\cite{Methods-2013-Hale}, 
seismic image is first computed before the estimation of the geologic fault location. Despite the success of those methods, 
there are  limitations. The success of the prediction heavily relies on the resulting seismic image or model obtained 
from the data sets. To avoid this limitation, another type of learning method has been recently proposed and developed, i.e.,
learning from pre-stack seismic data directly. In the work of \cite{Araya-2017-Automated} and \cite{Zhang-2014-Machine}, supervised learning
methods are directly applied to the pre-stack seismic data to look for patterns which indicates the geologic features. Specifically, 
in~\cite{Araya-2017-Automated} deep neural network was utilized to seismic data sets to obtain geologic faults. 
In~\cite{Zhang-2014-Machine}, kernel regression was used to learn a mapping between seismic data and geologic faults. In 
both paper, promising results have been reported.

In this work, our novel geologic feature detection is  belong to the learning-from-data category, meaning our algorithm is to detect
geological features from pre-stack seismic data sets.  
Through our experiments, we notice that despite of the success of those existing learning-from-data methods in 
controlled experiments, they are limited by their computational efficiency, mostly due to the need to process large 
volumes of high-dimensional data. Consequently, none of the existing solutions are suitable for real-time or even near 
real-time detection. In typical exploratory geophysics applications, strongly rectangular data arise, which implies 
that the number of receivers is much smaller than the number of data points that each receiver collects. Hence, we 
develop a scalable geologic feature detection technique by utilizing tools from randomized linear algebra allowing 
computational efficient geological feature detection.

Randomized matrix approximation methods enable us to efficiently deal with large-scale problems by sacrificing a provably 
trivial amount of accuracy~\citep{Drineas-2016-RandNLA}. Broadly, the underlying idea is to perform dimensionality reduction 
on the large-scale matrix without losing information pertinent to task under consideration. Their main idea is to construct 
a sketch matrix of the input matrix. The sketch matrix is usually a smaller matrix that yields a good approximation and 
represents the essential information of the original input. In essence, a sketching matrix is applied to the data to obtain 
a sketch that can be employed as a surrogate for the original data to compute quantities of interest~\citep{Drineas-2016-RandNLA}.
Randomized algorithms have been successfully applied to various scientific and engineering domains, such as scientific computation 
and numerical linear algebra~\citep{Meng-2014-LSRN, Drineas-2011-Fast, lin-2010-upre, Rokhlin-2008-Fast}, 
seismic full-waveform inversion and tomography~\citep{Moghaddam-2013-New, Krebs-2009-Fast}, and medical 
imaging~\citep{Huang-2016-Breast, Wang-2015-Waveform, Zhang-2012-Efficient}.

In this paper, we developed a novel geologic feature detection methods based on randomization. In particular, we consider 
the use of kernel machines for automated feature detection and design a scalable algorithm 
using the Nystr\"om approximation~\citep{Drineas-2005-Approximating, gittens2013revisiting}. 
It is well known that the kernel 
matrix used in any kernel machines is the bottleneck for scaling up computation and memory efficiency. The main idea of 
Nystr\"om method is to approximate an arbitrary symmetric positive semidefinite (SPSD) kernel matrix using a small subset of 
its columns, and the method reduces the time complexity of many matrix operations from $\mathcal{O}(n^2)$ or $O (n^3)$ to $\mathcal{O}(n)$
and space complexity from $\mathcal{O}(n^2)$ to $\mathcal{O}(n)$, where $n$ is the number of data samples.
There has been various research work utlizing Nystr\"om approximation to improve
the computational efficiency and memory usage in machine learning community. \cite{williams2001using} used the  Nystr\"om  method 
to speedup matrix inverse such that the inference of large-scale Gaussian process regression can be efficiently performed. 
Later on, the  Nystr\"om  method has been applied to spectral clustering~\citep{Li-2011-Time, Spectral-2004-Fowlkes}, kernel 
SVMs~\citep{Zhang-2008-Improved}, and kernel PCA and manifold  learning~\citep{Large-2013-Talwalkar}, etc. In this work, we 
employ Nystr\"om approximation to kernel ridge regression. Instead of forming the full kernel matrix from seismic data sets, 
we generate a low-rank approximation of the full kernel matrix by using Nystr\"om approximation. We further validate the 
performance of our new subsurface geologic feature detection method using synthetic surface seismic data. Our proposed 
detection method significantly improves the computational efficiency while maintaining the accuracy of the full model. 

In the following sections, we first briefly describe some fundamentals of underlying geology and the governing physics of 
our problem of interests. We then go through the data driven approaches - kernel ridge regression model~(Sec.~\ref{section:Theory}). 
We develop and discuss our novel geologic feature detection method based on randomized kernel ridge regression 
method~(Sec.~\ref{sec:DataReduction}). 
We then apply our method to test problems using both acoustic and elastic velocity models, and further discuss the results~(Sec.~\ref{sec:Results}). 
Finally, concluding remarks are presented in Sec.~\ref{sec:Conclusions}.

\section{Theory}
\label{section:Theory}

\subsection{Geologic Features of Interest: Fault Zones}

Geologic fault zone provides critical information for various subsurface energy applications. As an example, in carbon 
sequestration, leakage of $\mathrm{CO}_2$ and brine along faults at carbon sequestration sites isa primary concern for 
stroage integrity~\citep{Probability-2009-Zhang}. Accurately siting the geologic fault zones is essential to monitor
the $\mathrm{CO}_2$ storage. We first provide some fundamentals on the geological fault. 

\begin{figure}
	\begin{center}
		\centering
		\includegraphics[width=0.35\columnwidth]{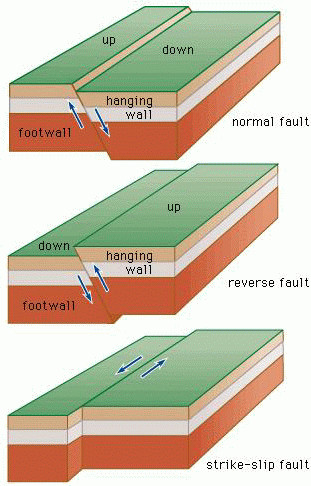}
	\end{center}
	\caption{An illustration of the geologic fault zones. There are three major geological fault types depending on 
                 the relative direction of displacement between the rocks on either side of the fault: normal fault, reverse 
                 fault, and strike-slip fault. The fault block above the fault surface is called the hanging wall, while the 
                 fault block below the fault is the footwall.}
	\label{fig:FaultIllustration}
\end{figure}

Geological fault is a fracture or crack along which two blocks of rock slide past one another~\citep{Haakon-2010-Structural}. 
As illustrated in Fig.~\ref{fig:FaultIllustration}, there are three major geological fault types depending on the relative 
direction of displacement between the rocks on either side of the fault: normal fault, reverse fault, and strike-slip fault. 
The fault block above the fault surface is called the hanging wall, while the fault block below the fault is the footwall.
In this study, we focus on both normal faults and reverse faults, which are the most common fault types~\citep{Haakon-2010-Structural}.

Out of various geophysical exploration methods, seismic waves are more sensitive to the acoustic/elastic impedance~(which depends on density and seismic velocity of the medium) of the subsurface than other geophysical measurements~(Fig.~\ref{fig:ExplorationGeophysics}). 
Hence, seismic exploration has been widely used to infer changes in the media impedance, which indicates geologic structures. 
In the next section, we briefly cover the mathematics and governing physics of seismic exploration.

\subsection{Physics-Driven Methods}
Seismic waves are mechanical perturbations that travel in the Earth at a speed governed by the acoustic/elastic impedance of the medium 
in which they are traveling. In the time-domain, the acoustic-wave equation is given by 
\begin{equation}
 \left [ \frac{1}{K(\mathbf{r})} \frac{\partial ^2}{\partial t ^2} 
        - \nabla  \cdot \left ( \frac{1}{\rho (\mathbf{r})}\,\, \nabla \right 
        ) \right ]
        p(\mathbf{r}, t) = s(\mathbf{r},\, t),
\label{eq:Forward}
\end{equation}
where $\rho (\mathbf{r})$ is the density at spatial location 
$\mathbf{r}$, $K(\mathbf{r})$ is the bulk modulus, $s(\mathbf{r},\, 
t)$ is the source term, $p(\mathbf{r}, t)$ is the pressure wavefield, 
and $t$ represents time.

The elastic-wave equation is written as
\begin{equation}
 \rho(\mathbf{r})\, \ddot{u}(r, t) - \nabla \cdot [C(\mathbf{r}) 
 : \nabla u(\mathbf{r}, t)] = s(\mathbf{r},\, t),
 \label{eq:ForwardElastic}
\end{equation}
where $C(\mathbf{r})$ is the elastic tensor, and $u(\mathbf{r}, t)$ is
the displacement wavefield.

The forward modeling problems in Eqs.~\eqref{eq:Forward} and 
\eqref{eq:ForwardElastic} can be written as
\begin{equation}
 P = f(\mathbf{m}),
 \label{eq:ForwardLinearM}
\end{equation}
where  $P$ is the pressure 
wavefield for the acoustic case or the displacement wavefields for the elastic case, $f$ is the forward acoustic or 
elastic-wave modeling operator, and $\mathbf{m}$ is the velocity model parameter vector, including the density and 
compressional- and shear-wave velocities.  We use a time-domain stagger-grid finite-difference scheme to solve the 
acoustic- or elastic-wave equation. Throughout this paper, we consider only constant density acoustic or elastic media. 

The inverse problem of Eq.~(\ref{eq:ForwardLinearM}) is usually posed 
as a minimization problem
\begin{equation}
E(\mathbf{m}) = \underset{\mathbf{m}}{\operatorname{min}} \left 
\{\left \| d - f(\mathbf{m})\right \| _2 ^2 + \lambda\, R(\mathbf{m}) 
\right \},
\label{eq:MisFit}
\end{equation}
where $\mathbf{d}$ represents a recorded/field waveform dataset, 
$f(\mathbf{m})$ is the corresponding forward modeling result, $ \left \| \mathbf{d} - f(\mathbf{m})\right \| _2 ^2$ is the data misfit,  
$||\cdot ||_2$ stands for the $\text{L}_2$ norm, $\lambda$ is a regularization parameter and $R(\mathbf{m})$ is 
the regularization term whose form depends on the type of the regularization used. 
The current technology to infer the subsurface geologic features is based on seismic inversion and imaging methods, which 
are computationally expensive and often yield unsatisfactory resolution in identifying small geologic 
features~\citep{Acoustic-2015-Lin, Quantifying-2015-Lin}. Recent years, with the significantly improved computational
power, machine learning and data mining have been successfully employed to a various domains from science to engineering. 
In the next section, we provide a different perspective (data driven approach) of extracting subsurface geological features from 
seismic measurements.

\subsection{Data-Driven Approach for Subsurface Feature Detection}

In this paper, we adopt a data-driven approach, which falls into the category of supervised machine learning.
Suppose one has $n$ historical \emph{features vectors} 
\begin{equation}
\mathbf{X} = [ \mathbf{x}_1^T, \cdots , \mathbf{x}_n^T ] ^T \in \RB^{n \times d},
\label{eq:features}
\end{equation}
which are from seismic measurements and $\mathbf{x}_i \in  \RB^{d \times 1}$,
and the associated \emph{labels} 
\begin{equation}
\mathbf{y} = [ y_1 , \cdots , y_n ] ^T \in \RB^{n \times 1},
\label{eq:labels}
\end{equation}
which can be the location or angle of geologic faults.

Overall, the idea of data driven approach independent of applications can be illustrated as
  \begin{equation*}
	\displaystyle	\mathrm{\textbf{Physical Measurements}}  \xrightarrow[]{f^\star} \mathrm{\textbf{Labels}} .
  \end{equation*}
In particular, one can build a machine learning model, such as kernel ridge regression (KRR), and \emph{train} the model using the recorded physical measurements.
After training, one gets a function, $f^\star$, which takes a $d$-dimensional feature vector as input and returns a prediction of its label.
Then for any unseen feature vector $\mathbf{x'} \in \RB^d$, one can predict its label by $f^\star (\mathbf{x'})$. 

\begin{figure*}
	\centerline{
		\subfloat[]{\label{fig:ExplorationGeophysics}
			\includegraphics[width=0.48\textwidth]{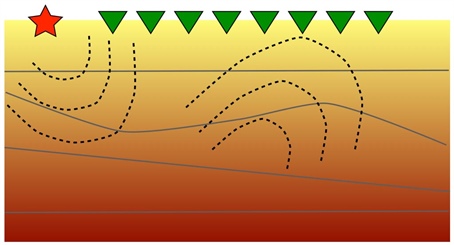}}
		\quad
		\subfloat[]{\label{fig:LearningProcedure-Measurements-cut}
			\includegraphics[width=0.48\textwidth]{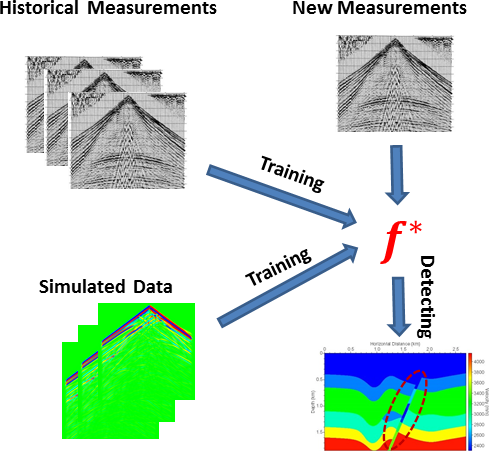}}}
	\caption{(a). An illustration of subsurface properties exploration by using seismic wave. (b). The diagram  of the data-driven procedure to learn geologic features from seismic measurements.  Seismic measurements including both historical and simulated are utilized as training data sets, which are fed into the learning algorithms. A mapping function, $f^\star (\mathbf{x'})$, 
is the outcome of the training algorithms. The function, $f^\star (\mathbf{x'})$, is the detection function, which creates a link from the seismic measurements to the corresponding 
geological features.}
	\label{fig:ExplorationFault}
\end{figure*}

As for subsurface geological feature detection specifically, we illustrate our data-driven approach in Fig.~\ref{fig:LearningProcedure-Measurements-cut}.
Seismic measurements including both historical and simulated are utilized as training data sets, which are fed into the learning algorithms. A mapping function, $f^\star (\mathbf{x'})$, 
is the outcome of the training algorithms. The function, $f^\star (\mathbf{x'})$, is the detection function, which creates a link from the seismic measurements to the corresponding 
geological features.

Note the difference between a machine learning model and the physical driven models as in Eq.~\eqref{eq:MisFit}.
A data driven model, such as KRR, is generic: it can be used to predict wine quality, web page click, house price, etc. 
To apply a data driven model, one need zero knowledge of the physics behind the problem;
one just need to provide the historical feature vectors and labels for training.
This is in sharp contrast to the physics driven model in Eq.~\eqref{eq:MisFit}, which is specific to one particular problem and requires strong domain knowledge and intricate mathematical models.

The correctness of our applied data-driven approach, KRR, is ensured by machine learning theory~\citep{friedman2001elements,mohri2012foundations}.
Assume that the training and test data are generated by the same model (otherwise, what is learned from the training data does not apply to the test data).
As more data are used for training, the prediction error monotonically decreases.
Importantly, KRR is known to be robust to noise: even if the training data are corrupted by intensive noise, the prediction is still highly accurate, provided that the number of training data is sufficiently large.
The robustness is useful in practice, because the seismic measurements have noise, and the locations and angles of the geologic faults may not be exactly known.

\subsection{Ridge Regression and Kernel Trick}

This work proposes to learn the function in question (denote $f^\star$) using data driven techniques such as ridge regression and kernel ridge regression~(KRR). 
Since all these regression methods are central to the proposed system, we recap their definitions in the following sections. 

\subsubsection{Ridge Regression}

Ridge regression is one of most popular regression methods, which models the linear dependencies between features vectors $\mathbf{x}$ and response variables $\mathbf{y}$.
Its loss function is usually posed as
\begin{equation}\label{eq:RidgeRegression}
\min_{\mathbf{w}} \bigg\{ \frac{1}{2} \sum_{i=1} ^{n} \big\|y_i - \mathbf{w}^T \mathbf{x}_i \big\|_2^2 + \frac{\lambda}{2}   \big\|  \mathbf{w}  \big\|^2  \bigg\},
\end{equation}
where the first term is the cost function and the second term is used to avoid the over fitting. The resulting regression model can be therefore obtained by 
\begin{equation}\label{eq:linearModel}
f^\star(x)  = \langle \mathbf{w}^T, \mathbf{x} \rangle,
\end{equation}
where $\mathbf{w}  =  (\mathbf{X} ^T \mathbf{X} + \lambda \mathbf{I_d}) ^{-1} \mathbf{X} ^T \mathbf{y}$. The major shortcoming of ridge regression is its limitation in modeling nonlinear data sets. However, in seismic applications, the relationship between the feature vectors and the response variables is nonlinear because of the governing physics as provided in Eqs.~\eqref{eq:Forward} and \eqref{eq:ForwardElastic}. We will need more advanced regression techniques to model the data nonlinearity while maintaining feasible computational costs. Kernel tricks provide us 
with the tools~\citep{scholkopf2002learning}. 

\subsubsection{Kernel Ridge Regression}
Kernel ridge regression (KRR) is a popular non-linear regression method \citep{scholkopf2002learning}. It is built upon kernel function $\kappa()$, which is defined as
\begin{definition} 
A function $\kappa: \RB^d \times
\RB^d \rightarrow \RB$ is a valid kernel if $\displaystyle \sum_{i=1}^n \sum_{j=1}^n z_i z_j \kappa (\mathbf{x}_i , \mathbf{x}_j) \geq 0$ holds  for any $\mathbf{x}_1, \cdots , \mathbf{x}_n \in \RB^{d}$ and $z_1 , \cdots , z_n \in \RB$. 
In addition, a valid kernel defines such a feature map $\phi: \RB^d \mapsto \FM$ that $\kappa(\mathbf{x}_i,\mathbf{x}_j) = \langle \phi(\mathbf{x}_i),\phi(\mathbf{x}_j) \rangle$, 
where $\langle \cdot,\cdot\rangle$ denotes the inner product in the \emph{feature space} $\FM$. 
\end{definition}
In general, a kernel $\kappa(\mathbf{x}_i, \mathbf{x}_j)$ measures the similarity between the two samples in $\RB^d$.
In this paper, we consider the radial basis function~(RBF) as the kernel which is defined as
\begin{equation} \label{eq:rbf}
\kappa(\mathbf{x}_i, \mathbf{x}_j) \; = \; \exp  \big( - \tfrac{1}{2\sigma^2} \|\mathbf{x}_i-\mathbf{x}_j \|_2^2  \big),
\end{equation} where $\sigma > 0$ is the kernel width parameter
and $\| \cdot \|_2$ is the vector Euclidean norm ($\ell_2$-norm). 
Since linear operations within the feature space $\FM$ can be interpreted as non-linear operations in the Euclidean space, the linear models learned in the feature space provide the power of a non-linear model. 
Let $K \in \RB^{n\times n}$ be the (training) kernel matrix,
where $k_{ij} = \kappa (\mathbf{x}_i, \mathbf{x}_j)$.

Suppose the seismic measurements are stored as
$(\mathbf{x}_1, y_1), \cdots , (\mathbf{x}_n , y_n) \in \RB^d \times \RB$.
KRR uses the data for training and returns a function $\hat{f}$ which approximates $f^\star$.
Given a test point $\mathbf{x'} \in \RB^d$, KRR makes prediction $\hat{f} (\mathbf{x'})$.
We directly state the dual problem of KRR without derivation;
the reader can refer to \citet{campbell2001introduction} for the details:
\begin{equation}\label{eq:KernelRidgeRegression}
\min_{\boldsymbol \alpha} \bigg\{ \frac{1}{2} \sum_{i=1} ^{n} \big\|y_i - (K\boldsymbol \alpha)_i \big\|_2^2 + \frac{\lambda}{2} \boldsymbol \alpha^T\,K\,\boldsymbol \alpha  \bigg\},
\end{equation}
where $\lambda > 0$ is the regularization parameter and should be fine tuned.
Problem~\eqref{eq:KernelRidgeRegression} has a closed-form optimal solution 
\begin{equation}\label{eq:krr_optimal}
\boldsymbol \alpha^\star 
\; = \; (K + \lambda I_n)^{-1} y 
\; \in \; \RB^n ,
\end{equation}
where $I_n$ is the $n \times n$ identity matrix.
Finally, for any $\mathbf{x'} \in \RB^d$, 
\begin{equation}
\label{eq:solutiontoKRR}
\hat{f} (\mathbf{x'}) \; = \; \sum_{i=1}^n \boldsymbol \alpha_i^\star \kappa (\mathbf{x'}, \mathbf{x}_i)
\end{equation}
is the prediction made by KRR.

Machine learning theory indicates that more training samples lead to smaller variance and thereby better prediction performance.
Ideally, one can collect as many seismic measurements as desired in the quest to improve detection. Unfortunately, the $\OM (n^3)$ time and $\OM (n^2)$ memory costs of KRR hinder the use of such large amounts of training data. To the best of our knowledge, these computational challenges of KRR have not been addressed by any of the prior efforts on using kernel machines for subsurface applications~\citep{Schnetzler-2017-Use, Zhang-2014-Machine, Ramirez-2011-Machine}. 
A practical approach to large-scale KRR is randomized kernel approximation, which sacrifices a limited amount of accuracy for a tremendous reduction in time and memory costs.
In this work, we apply the Nystr\"om method \citep{nystrom1930praktische,williams2001using} to make large-scale KRR feasible on a single workstation.
Consequently, we can easily enable the training of KRR using much larger amounts seismic measurements, thereby achieving substantially improved geologic detection performance.

To estimate the subsurface features using seismic data sets, the number of samples, $n$, is usually in the magnitude of millions or even more. 
KRR requires forming an $n\times n$ kernel matrix $K$ in Eq.~\eqref{eq:KernelRidgeRegression} and inverting the $n\times n$ matrix in \eqref{eq:krr_optimal}.
Therefore, when $n$ is large, the computation of KRR is very challenging. Super computer clusters and distributed computing systems have been utlized to solve Large-scale KRR~ \citep{avron2017faster,tu2016large}. Even though KRR provides promising regression results in various problems, such an expensive computational cost of KRR will 
unfortunately hinders its wide  applications. A practical approach to large-scale KRR is randomized kernel approxation, which sacrifices a limited amount of accuracy for a tremendous reduction in time and memory costs.
We apply the Nystr\"om method \citep{nystrom1930praktische,williams2001using} to make large-scale KRR feasible on a single workstation.

\section{Scalable Geologic Detection through Randomized Approximation}
\label{sec:DataReduction}

In this section, we introduce the Nystr\"om method---a randomized kernel matrix approximation tool---to the geologic detection task, aiming at solving large-scale problems using limited computational resources.
Sec.~\ref{sec:DefineNystrom} describes the Nystr\"om method,
Sec.~\ref{sec:SelectionOFsketching} theoretically justifies the Nystr\"om method and its application to KRR,
Sec.~\ref{sec:TuningParam} discusses the three tuning parameters,
Sec.~\ref{sec:OverallAlgorithm} presents the whole procedure of KRR with Nystr\"om approximation,
and finally, Sec.~\ref{sec:CompMemoryCost} analyzes the time and memory costs.

\subsection{The Nystr\"om Method}
\label{sec:DefineNystrom}

Nystr\"om approximation \citep{williams2001using} is a popular and an efficient approach.
In addition to its simplicity, the Nystr\"om method is a theoretically sound approach:
its approximation error is bounded \citep{Drineas-2005-Approximating,gittens2013revisiting};
when applied to KRR, its statistical risk is guaranteed to diminish\citep{alaoui2015fast,bach2013sharp}.

\begin{figure}
	\begin{center}
		\centering
		\includegraphics[width=0.95\columnwidth]{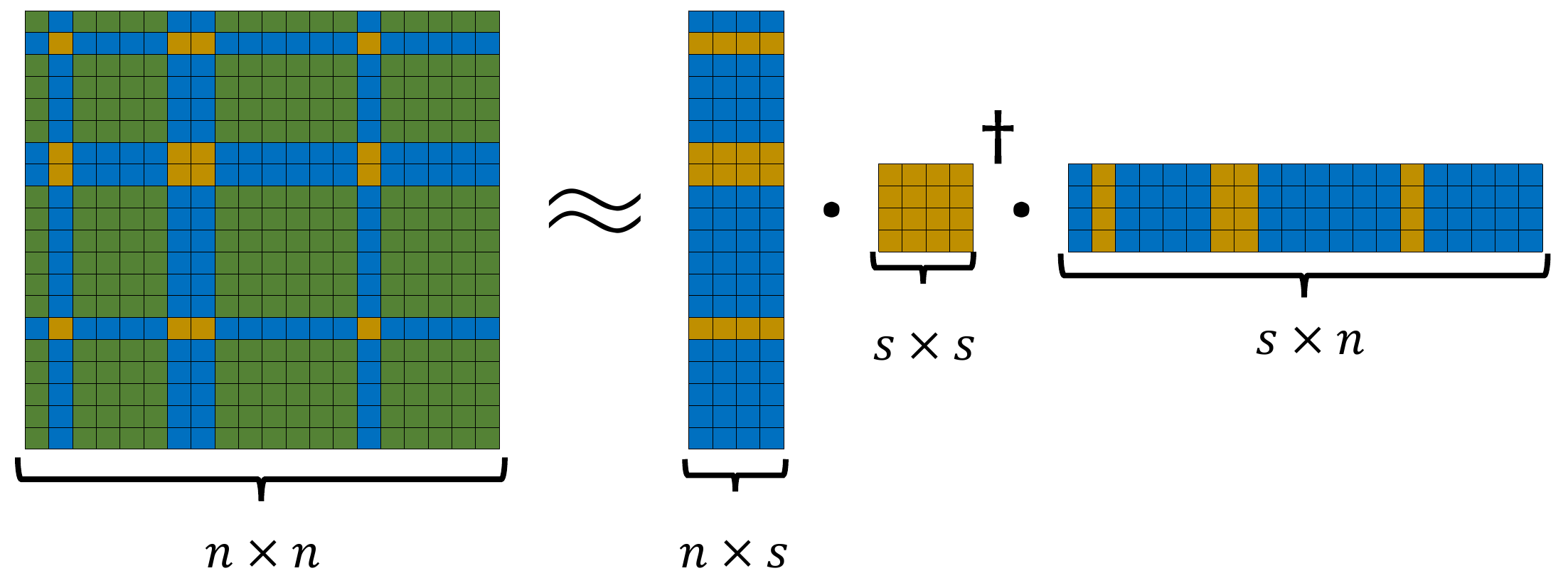}
	\end{center}
	\caption{Illustration of the Nystr\"om approximation $K \approx \Psi \Psi^T  = C W^\dag C^T$, where the low-rank approximation $\Psi = C (W^\dagger)^{1/2}$ can be obtained. }
	\label{fig:nystrom}
\end{figure}

The Nystr\"om method computes a low-rank approximation $K \approx \Psi \Psi^T \in \RB^{n\times n}$ in $\OM (n d s + n s^2)$ time.
Here $s \ll n$ is user-specified; larger values of $s$ leads to better approximation but incurs higher computational costs.
The tall-and-skinny matrix $\Psi \in \RB^{n\times s}$ is computed as follows:
First, sample $s$ items from $\{1, \cdots , n\}$ uniformly at random without replacement;
let the resulting set be $\SM$.
Subsequently, construct a matrix $C \in \RB^{n\times s}$ as $c_{il} = \kappa (\mathbf{x}_i, \mathbf{x}_l)$ for $i \in \{1, \cdots , n\}$
and $l \in \SM$;
let $W \in \RB^{s\times s}$ contain the rows of $C$ indexed by $\SM$.
\citet{gittens2013revisiting} showed that $C W^\dag C^T$ is a good approximation to $K$,
where $W^\dag$ denotes the Moore-Penrose pseudo-inverse of $W$.
The approximation is illustrated in Fig.~\ref{fig:nystrom}.
Finally, the low-rank approximation $\Psi = C (W^\dagger)^{1/2}$ is computed.

Besides the Nystr\"om method, a number of other kernel approximation methods exist in the machine learning literature.
Random feature mapping \citep{le2013fastfood,rahimi2007random} is an equally popular class of approaches.
However, compared to random feature mapping, several theoretical and empirical studies \citep{tu2016large,yang2012nystrom} are in favor of the Nystr\"om method.
Furthermore, in the recent years, alternative approaches such as the fast SPSD model \citep{wang2015towards}, MEKA~\citep{si2014memory}, 
hierarchically compositional kernels~\citep{chen2016hierarchically} have been proposed to speed up KRR.
Since comparing different kernel approximation methods is beyond the scope of this work,
we adopt the Nystr\"om method in our algorithm.

\subsection{Theoretical Justifications of the Nystr\"om Method}
\label{sec:SelectionOFsketching}

The Nystr\"om method has been studied by many recent works \citep{alaoui2015fast,bach2013sharp,Drineas-2005-Approximating,gittens2013revisiting,wang2015towards,wang2017scalable}, and its theoretical properties has been well understood.
In the following, we first intuitively explain why the Nystr\"om method works and then describe its theoretical properties.

Let $P \in \RB^{n\times s}$ be a column selection matrix, that is, each column of $P$ has exactly one nonzero entry whose position indicates the index of the selected column.
We let $K = K^{\frac{1}{2}} K^{\frac{1}{2}} $ and $D = K^{\frac{1}{2}} P$.
Then the matrices $C$ and $W$ (in Fig.~\ref{fig:nystrom}) can be written as
\[
C = K P = K^{\frac{1}{2}} K^{\frac{1}{2}} P = K^{\frac{1}{2}} D
\quad
\textrm{and}
\quad
W = P^T K P 
= P^T  K^{\frac{1}{2}} K^{\frac{1}{2}} P 
= D^T D.
\]
The Nystr\"om approximation can be written as
\[
K 
\: \approx \:
C W^\dag C^T
\: = \: 
K^{\frac{1}{2}} 
\big[ D (D^T D)^\dag D \big] 
K^{\frac{1}{2}} 
\]
In the extreme case where $s=n$, the matrix $D (D^T D)^\dag D$ is identity, and thus the Nystr\"om approximation is exact.
In general $s < n$, the matrix $D (D^T D)^\dag D$ is called orthogonal projection matrix, which projects any matrix to the column space of $D$.
Low-rank approximation theories show that if the ``information'' in $K$ is spread-out, then most mass of $K^{\frac{1}{2}} $ are in the column space of a small subset of columns of $K^{\frac{1}{2}} $.
Theorefore, projecting $K^{\frac{1}{2}} $ to the column space of $D = K^{\frac{1}{2}} P$ loses only a little accuracy, and the Nystr\"om approximation $K^{\frac{1}{2}} \big[ D (D^T D)^\dag D \big] K^{\frac{1}{2}}$ well approximates $K$.

Theoretical bounds \citep{gittens2013revisiting,wang2017scalable} guarantee that the Nystr\"om approximation $C W^\dag C^T$ is close to $K$ in terms of matrix norms.
Let $k$ ($\geq 1$) be arbitrary integer, $K_k$ be the best rank $k$ approximation to $K$, and $\| \cdot \|$ be the spectral norm, Frobenius norm, or trace norm.
If the eigenvalues of $K$ decays rapidly and the number of samples, $s$, is sufficiently larger than $k$,
then $\| K - C W^\dag C^T \|$ is comparable to $\| K - K_k \|$.

Applied to the KRR problem, the quality of the Nystr\"om method has been studied by \cite{alaoui2015fast,bach2013sharp}.
The works studied the bias and variance, which directly affect the prediction error of KRR.
The works showed that using the Nystr\"om approximation, the increases in the bias is bounded, and the variance does not increase at all.
Therefore, using the Nystr\"om approximation, the prediction made by KRR will not be much affected.
In addition, they showed that as the number of samples, $s$, increases, the performance monotonically improves.

\subsection{Tuning Parameters}
\label{sec:TuningParam}

KRR with Nystr\"om approximation has totally three parameters:
the regularization parameter $\lambda$, 
the kernel width parameter $\sigma$,
and the number of random samples $s$.
We discuss the effect of the parameters.

The regularization parameter $\lambda$ ($\geq 0$) is defined in the KRR objective function \eqref{eq:KernelRidgeRegression} and can be arbitrarily set by users.
From the statistical perspective, $\lambda$ trades off the bias and variance of KRR:
big $\lambda$ leads to small variance but big bias, and vice versa.
The optimal choice of $\lambda$ is the one minimizes the sum of variance and squared bias.
However, such optimal choice cannot be analytically calculated; in practice, it is determined by cross-validation.\footnote{Cross-validation is a standard machine learning technique for tuning parameters. One can randomly split the training set into two parts, train on one part, make prediction on the other, and choose the parameter corresponding to the best prediction error.}

The kernel width parameter $\sigma$ is defined in \eqref{eq:rbf}.
It defines how far the influence of a single training example reaches, with high values meaning ``far'' and high values meaning ``close''.
As $\sigma$ goes to zero, $K$ tends to be identity, where the training examples do not influence each other and the KRR model is too flexible;
as $\sigma$ goes to infinity, $K$ tends to be an all-one matrix (its rank is one), where the KRR model is restrictive and lacks expressive power.
The users can either set $\sigma$ as In practice, $\sigma$ should be fine tuned; a good heuristic is searching $\sigma$ around
\begin{equation} \label{eq:SetSigma}
\displaystyle \sqrt{\textstyle \frac{1}{n^2} \sum_{i=1}^n \sum_{j=1}^n \|\mathbf{x}_i - \mathbf{x}_j\|_2^2 } .
\end{equation} 
or searching $\sigma$ around this value by cross-validation.
Note that computing \eqref{eq:SetSigma} costs $\OM ( n^2 d)$ time and is thereby impractical,
a good heuristic is randomly sample a subset $\JM \subset \{1, \cdots , n \}$ and approximate \eqref{eq:SetSigma} by
\begin{equation*} 
\sqrt{\textstyle \frac{1}{|\JM |^2} \sum_{i \in \JM} \sum_{j \in \JM } \|\mathbf{x}_i - \mathbf{x}_j\|_2^2 } ,
\end{equation*} 
which costs merely $\OM ( d |\JM |^2)$ time.

The number of random samples $s$ trades off the accuracy and computational costs:
large $s$ leads to good prediction but large computational costs.
If the dataset has $n$ samples of $d$-dimension, the total time complexity is $\OM (n d s + n s^2)$,
and the space (memory) complexity is $\OM (nd + ns)$.
It is always good to set $s$ as large as one can afford because the prediction monotonically improves as $s$ increase.

\subsection{Overall Algorithm: KRR with Nystr\"om Approximation}
\label{sec:OverallAlgorithm}

Using the Nystr\"om method, the training of KRR can be performed in $O(n d s + n s^2)$ time,
where the user-specified parameter $s$  directly trades off accuracy and computational costs.
Empirically speaking, setting $s$ in the order of hundreds suffices in our application.
The overall algorithm is described as follows.

{\bf Training.}
The inputs are $(\mathbf{x}_1, y_1), \cdots , (\mathbf{x}_n, y_n) \in \RB^{d} \times \RB$.
User specifies $s$, randomly select $s$ out of the $n$ samples,
form the kernel sub-matrices $C \in \RB^{n\times s}$ and $W \in \RB^{s\times s}$,
and compute $\Psi = C W^{-\frac{1}{2}}$.
The kernel matrix $K$ can be approximated by $\Psi \Psi^T$ according to the previous subsection.
Finally, $\alpha^\star$ defined in Eq.~\eqref{eq:krr_optimal} can be approximated by
\begin{eqnarray} 
\hspace{4cm}\tilde{\boldsymbol \alpha}
\label{eq:TildeAlpha1}
& = & \big(\Psi \Psi^T + \lambda I_n \big)^{-1} y,  \\
\label{eq:TildeAlpha2}
& = & \lambda^{-1} y - \lambda^{-1} \Psi (\lambda I_s + \Psi^T \Psi )^{-1} \Psi^T y
\; \in \; \RB^n ,
\end{eqnarray}
where the latter equality follows from the Sherman-Morrison-Woodbury~(SMW) matrix identity as definied in Eq.~\eqref{eq:Woodbury} and the detailed deviration is provided from Eq.~\eqref{eq:SMW_Deviration1} to Eq.~\eqref{eq:SMW_Deviration2}. More details can be found in \cite{wang2015practical}. It is worthwhile 
to mention that the $n\times n$ matrix of $\Psi \Psi ^T$ in Eq.~\eqref{eq:TildeAlpha1} has been replaced by the  matrix of $\Psi^T \Psi$ in Eq.~\eqref{eq:TildeAlpha2}, which is a much smaller dimension of $s \times s$. This 
leads to the significant reduction of the computational costs.

{\bf Prediction.}
Let $\mathbf{x'} \in \RB^d$ be any unseen test sample.
The detection step is almost identical to Eq.~\eqref{eq:solutiontoKRR}:
we use $\tilde{\alpha} $ instead of $\alpha^\star$ and makes prediction by
\begin{equation*}
\hat{f} (x') 
\; = \; \sum_{i=1}^n {\boldsymbol \alpha}^\star_i \kappa (\mathbf{x'}, \mathbf{x}_i)
\; \approx \; \sum_{i=1}^n \tilde{\boldsymbol \alpha}_i \kappa (\mathbf{x'}, \mathbf{x}_i) .
\end{equation*}
The location or angle of geological fault should be close to $\hat{f} (\mathbf{x'}) $.

\subsection{Computational and Memory Cost Analysis}
\label{sec:CompMemoryCost}

The training of KRR without kernel approximation has $O (n^2 d + n^3)$ time complexity and $O (nd + n^2)$ space (memory) complexity.
The costs are calculated as follows. 
For most kernel functions, including the RBF kernel,
the evaluation of $\kappa (x_i, x_j)$ costs $\OM (d)$ time.
One needs to keep the $n$ data samples in memory to compute every entry of $K \in \RB^{n\times n}$,
which costs $\OM (n d)$ memory and $\OM (n^2 d)$ time.
To compute $\alpha^\star$, one needs to keep $K$ in memory and perform matrix inversion,
which requires $\OM (n^2)$ memory and $\OM (n^3)$ time.

The training of KRR with Nystr\"om approximation has $\OM (n d s + n s^2)$ time complexity and $\OM (n d + n s)$ space complexity.
The costs are calculated as follows. 
To compute $C \in \RB^{n\times s}$ and $W \in \RB^{s\times s}$, one only need to evaluate $n s$ kernel functions,
which requires $\OM (n d)$ memory and $\OM (n d s)$ time.
The computation of $\tilde{\alpha}$ according to Eq.~\eqref{eq:TildeAlpha2}
has $\OM (ns)$ space complexity (because the matrice $C$ and $W$ need to be kept in memory)
and $\OM (n s^2)$ time complexity.

The prediction of KRR, either with or without approximation, for an unseen test sample, $\mathbf{x'}$, has $\OM (nd)$ time complexity and $\OM(n d)$ memory complexity.
First, one keeps the $n$ data samples in memory to evaluate the kernel functions $\kappa(\mathbf{x'}, \mathbf{x}_1) , \cdots , \kappa (\mathbf{x'}, \mathbf{x}_n)$, which costs $\OM (n d)$ time and $\OM (n d)$ memory.
Then, one keeps the $n$ kernel function values and $\boldsymbol \alpha^\star$ (or $\tilde{\boldsymbol \alpha}$) in memory to make prediction, which costs merely $\OM (n )$ time and $\OM (n )$ memory.

\section{Numerical Results}
\label{sec:Results}

\begin{figure*}
\centering
\includegraphics[width=.45\linewidth]{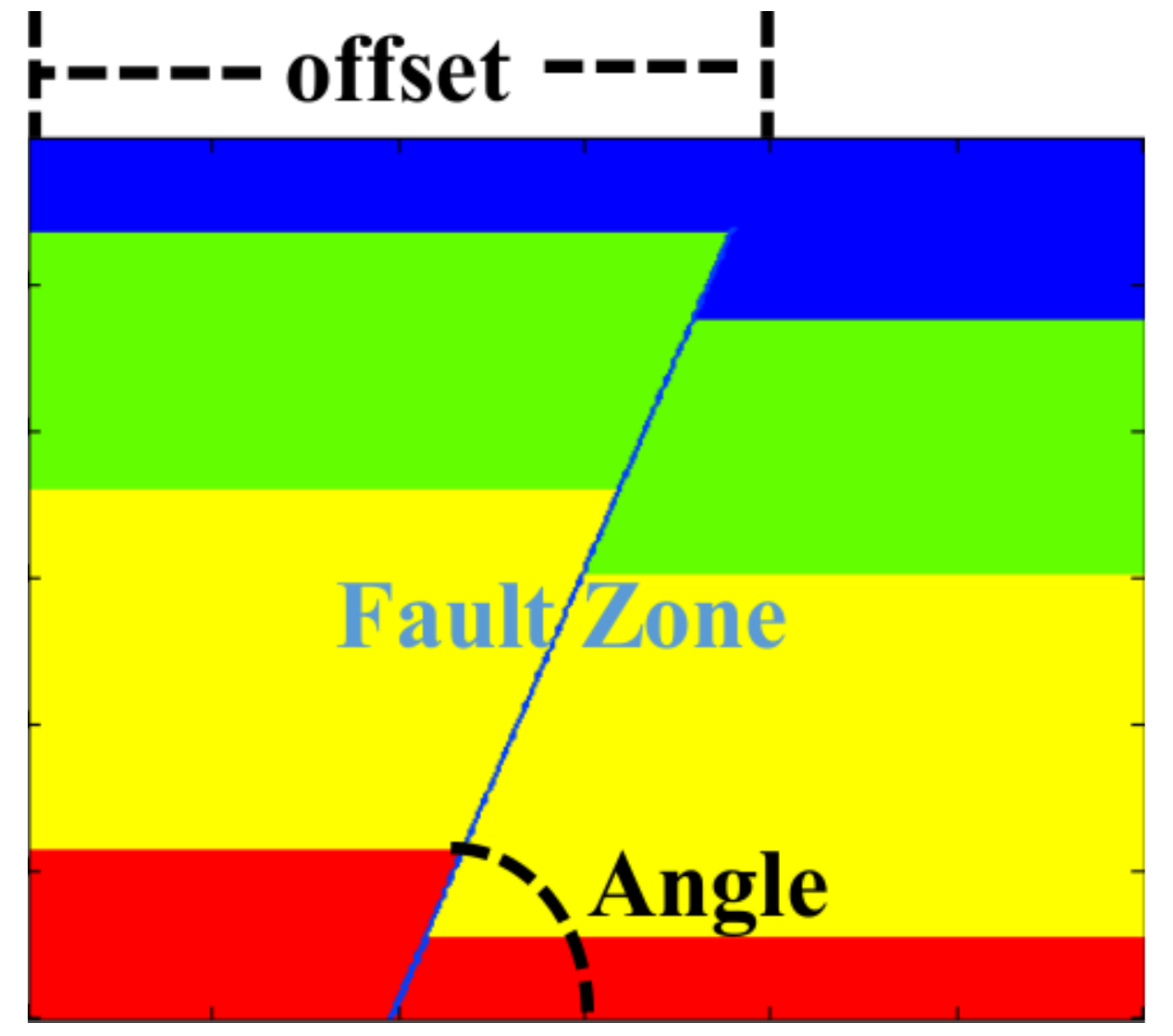}
  \caption{An illustration of the geologic fault zone. The location of a geologic fault zone can be characterized by its horizontal 
offset and the dipping angle~\citep{Zhang-2014-Machine}. }
\label{fig:GeologicFaultModel}
\end{figure*}

To validate the performance of our proposed approach, we carry out evaluations with synthetic seismic measurements to 
detect the location of the geologic faults. The siting of geologic fault zones can be characterized by its horizontal 
offset and the dipping angle as shown in Fig.~\ref{fig:GeologicFaultModel}~\citep{Zhang-2014-Machine}. We employ our 
new subsurface feature detection method to estimate both the offset and angle of geologic fault zones. As for the 
computing environment, we run our tests on a computer with 40 Intel Xeon E5-2650 cores running at 2.3~GHz, and 64~GB memory.

\begin{figure*}
\centering
\includegraphics[width=.60\linewidth]{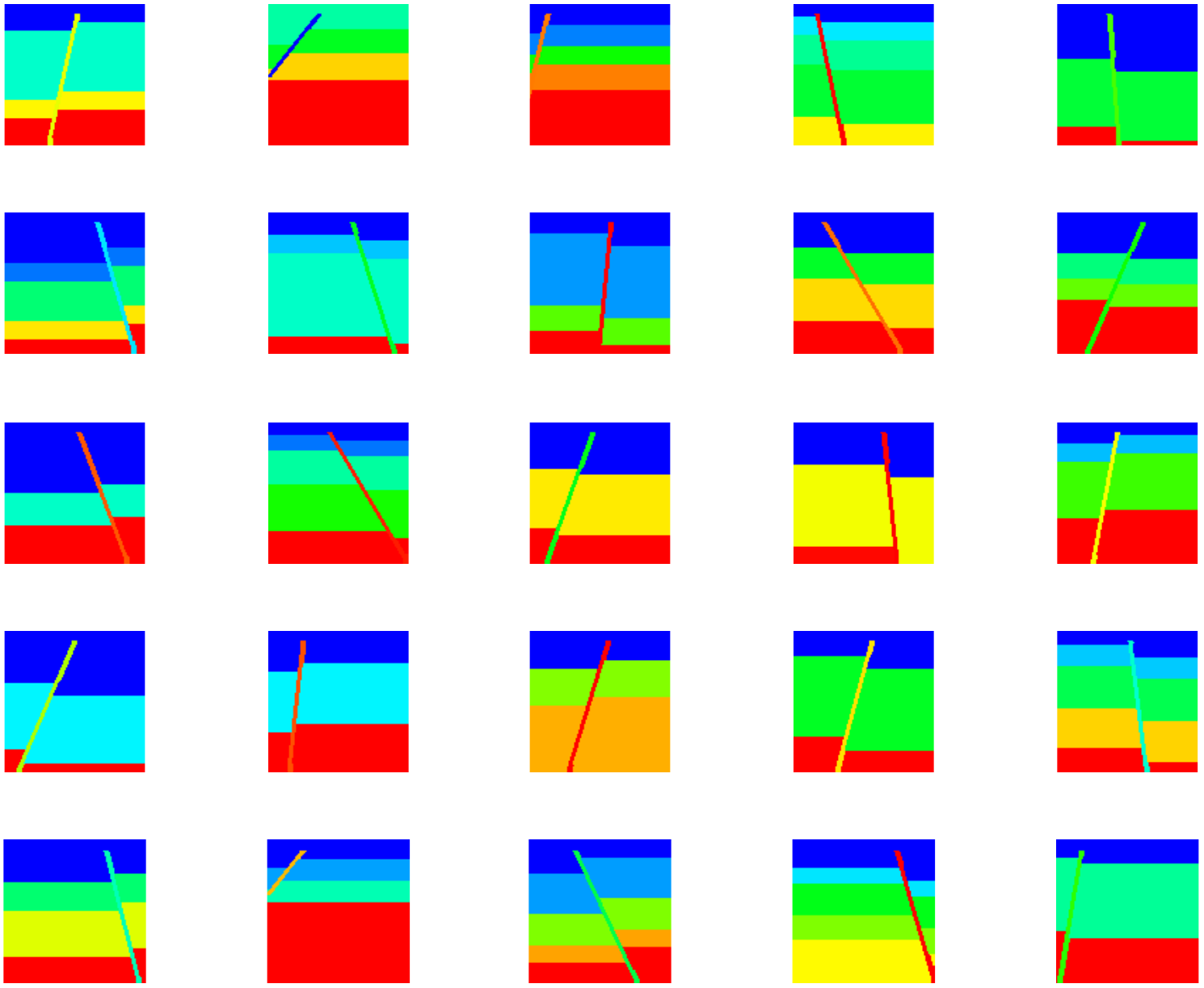}
\caption{A database of velocity models consisting of $60,000$ models of size $100 \times 100$ grid points.
         The velocity models in the database are different from one another in terms of 
         offset~(ranging from 30 grids to 70 grids), dipping angle~(ranging from $25^{\circ}$ to $165^{\circ}$), 
         number of layers~(ranging from 3 to 5 layers), layer thickness~(ranging from 5 grids to 80 grids), 
         and layer velocity~(ranging from 3000~m/s to 5000~m/s).}
\label{fig:TrainingSeismicModel}
\end{figure*}

The quality of training set is critical for any machine-learning algorithms. In this work, we consider velocity models 
consisting of horizontal reflectors with a single fault zone~(layer model) to demonstrate 
the performance of our new geologic feature detection method. It is straight forward to employ our method to 
detect multiple fault zones. To best represent the geologically realistic velocity models, we create a 
database containing $n = 60,000$ velocity models of size $100 \times 100$ grid points similar to~\cite{Zhang-2014-Machine}. 
The velocity models in the database are different from one another 
in terms of offset~(ranging from 30 grids to 70 grids), dipping angle~(ranging from $25^{\circ}$ to $165^{\circ}$), 
number of layers~(ranging from 3 to 5 layers), layer thickness~(ranging from 5 grids to 80 grids), and layer 
velocity~(ranging from 3000~m/s to 5000~m/s). A small portion of the training velocity models are shown in 
Fig.~\ref{fig:TrainingSeismicModel}.

\begin{figure*}
\centering
\subfloat[]{\includegraphics[width=.45\linewidth]{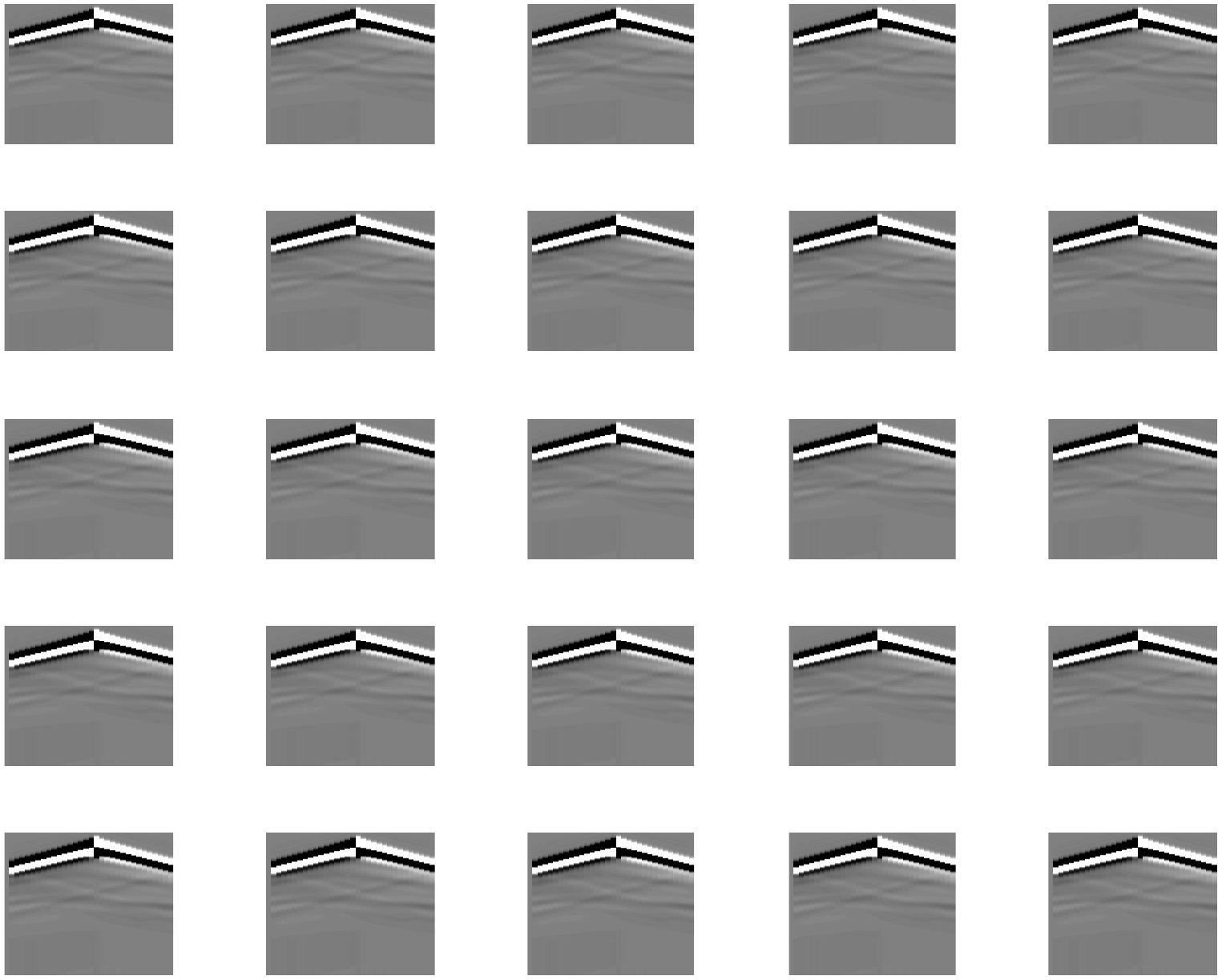}}
\quad
\quad
\quad
\subfloat[]{\includegraphics[width=.45\linewidth]{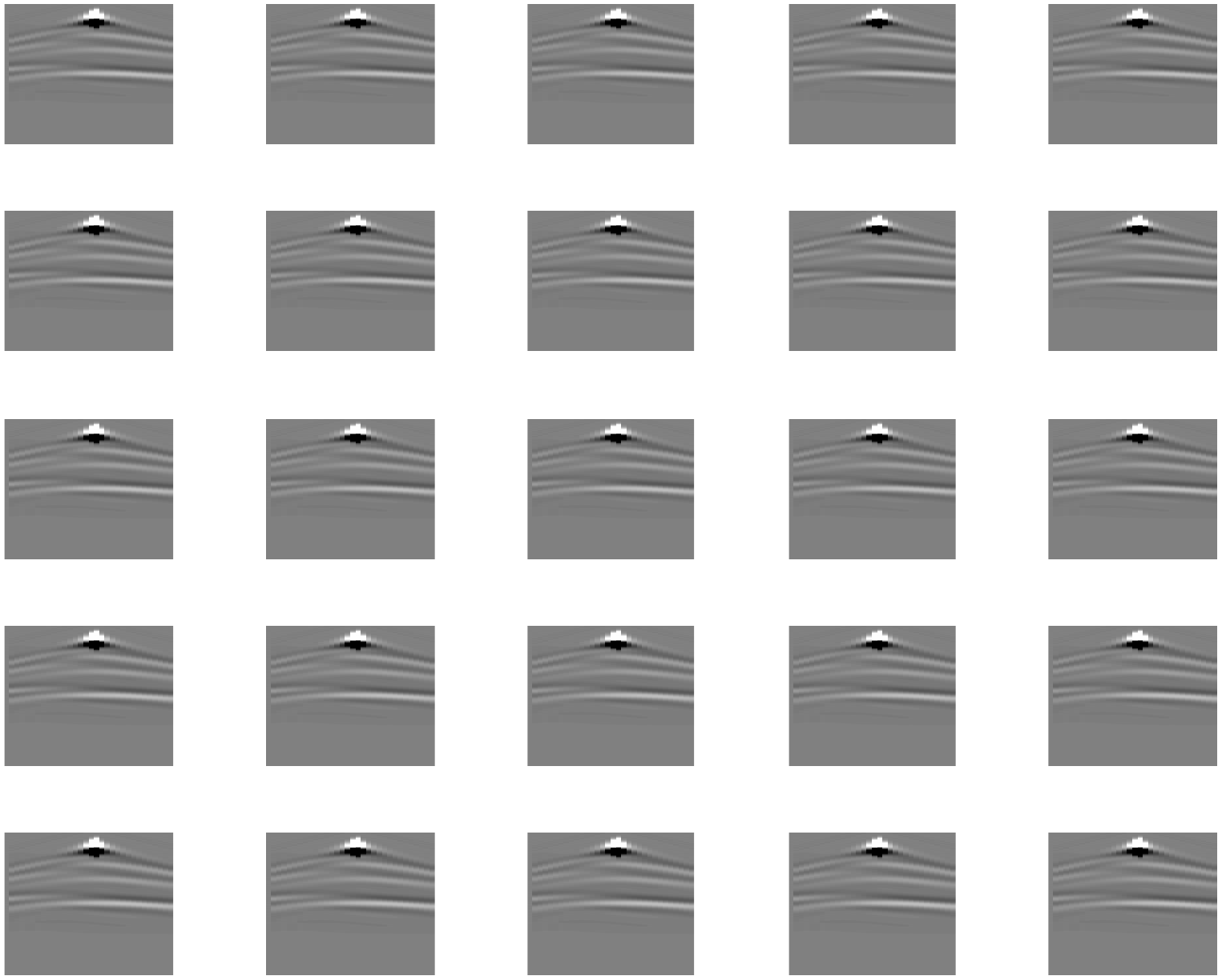}}
\caption{Synthetic seismic data sets are obtained using a staggered-grid finite-difference scheme with a perfectly matched 
         layered absorbing boundary condition. The displacement of X direction (a) and Z direction (b) are both used as 
         training sets. The total volume of synthetic seismic data is $1.92\times 10^{9}$.}
\label{fig:TrainingSeismicData}
\end{figure*}

The seismic measurements are collections of synthetic seismograms obtained by implementing forward modeling on those 
$60,000$ velocity models. One common-shot gather of synthetic seismic data with $32$ receivers is posed at the top surface 
of the model. The receiver interval is $15$~m. We use a Ricker wavelet with a center frequency of $25$~Hz as the source time 
function and a staggered-grid finite-difference scheme with a perfectly matched layered absorbing boundary condition to generate 
synthetic seismic reflection data. The synthetic trace at each receiver is a collection of time-series data of length $1,000$. So, the feature dimension of seismic data sets is $d = 3.2 \times 10^{4}$. Therefore, out of $60,000$ velocity models, the total volume of synthetic seismic data is $1.92\times 10^{9}$. In 
Fig.~\ref{fig:TrainingSeismicData}, we show a portion of the synthetic seismic data sets corresponding to velocity models that we generate. Specifically, the displacement in X direction is shown in Fig.~\ref{fig:TrainingSeismicData}(a), and the displacement in Z direction is shown in Fig.~\ref{fig:TrainingSeismicData}(b).

We employ a hold-out test to assess the efficacy of our proposed algorithm. Specifically, $75.0\%$ of the dataset is used for 
training the model, while the rest is used for testing. 
For comparison, we use the conventional KRR method (denoted by ``KRR'') as the reference method. 
We denote our new geologic feature detection method as ``R-KRR'' standing for Randomized KRR method. 
To evaluate the performance, we report both the accuracy
and the computational efficiency of different methods. We use the mean-absolute error~(MAE) metric to quantify the accuracy of a learning method, which
is defined as 
\begin{equation}
\displaystyle \mathrm{MAE}(\mathbf{y}, \hat{\mathbf{y}}) = \frac{1}{n} \sum_{i=1}^{n} |y_i - \hat{y_i}|.
\label{eq:MAE}
\end{equation}
 We calculate the wall time to measure the 
computational efficiency of a learning method and further provide the speed-up ratio. 

To have a comprehensive understanding our randomized geologic feature detection methods, we provide three sets of tests.
In Sec.~\ref{sec:TestonAccuracy}, we provide an overall test on the detection accuracy of our method.
In Sec.~\ref{sec:TestonRank}, we report the performance of our method as a function of the number of random samples, $s$. 
In Sec.~\ref{sec:TestonRandomness}, we test the robustness of our method with a view on the randomness of the approximation method.

\subsection{Test on Detection Accuracy}
\label{sec:TestonAccuracy}

\begin{figure*}
\centering
  \subfloat[]{\includegraphics[width=.50\linewidth]{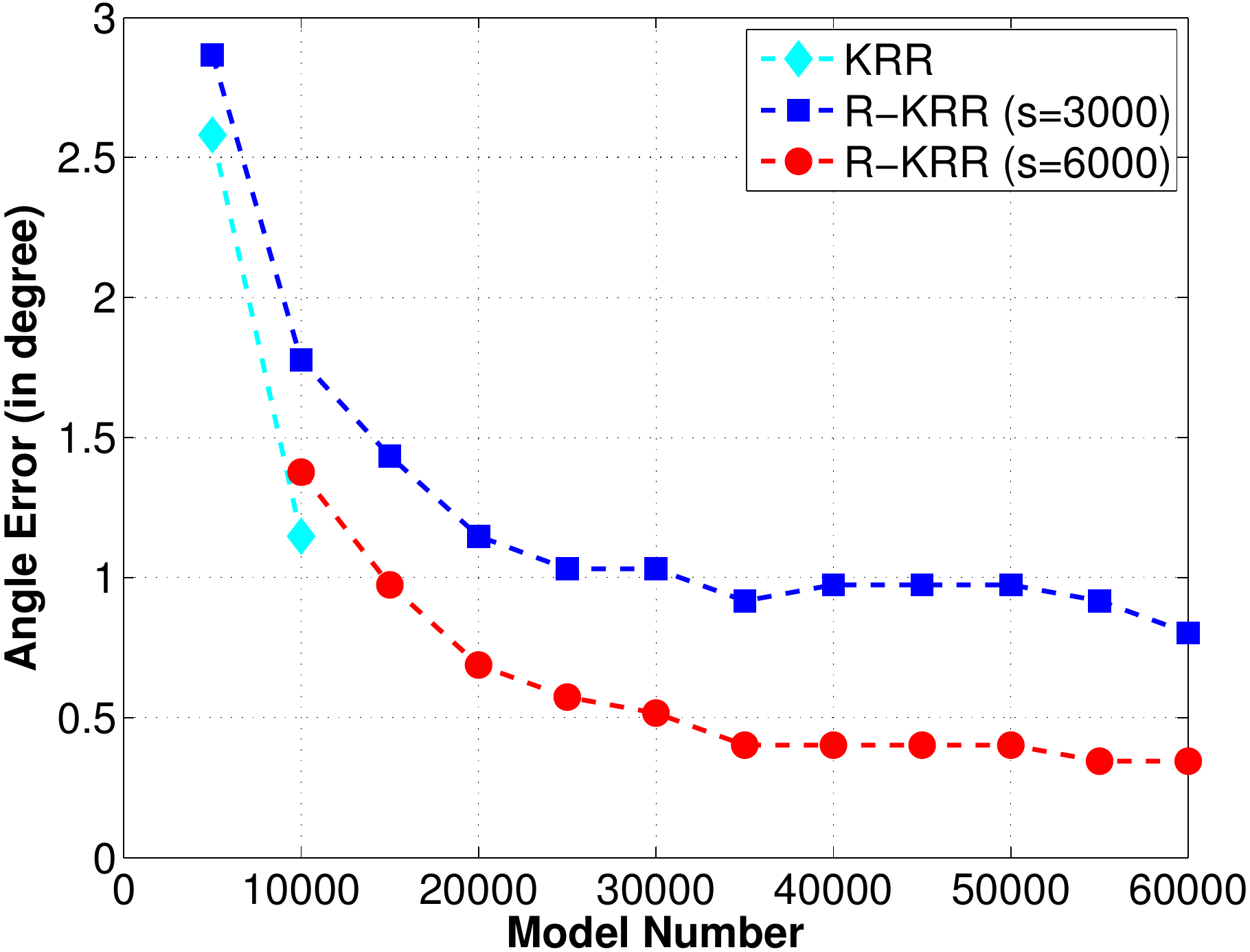}}
  \subfloat[]{\includegraphics[width=.50\linewidth]{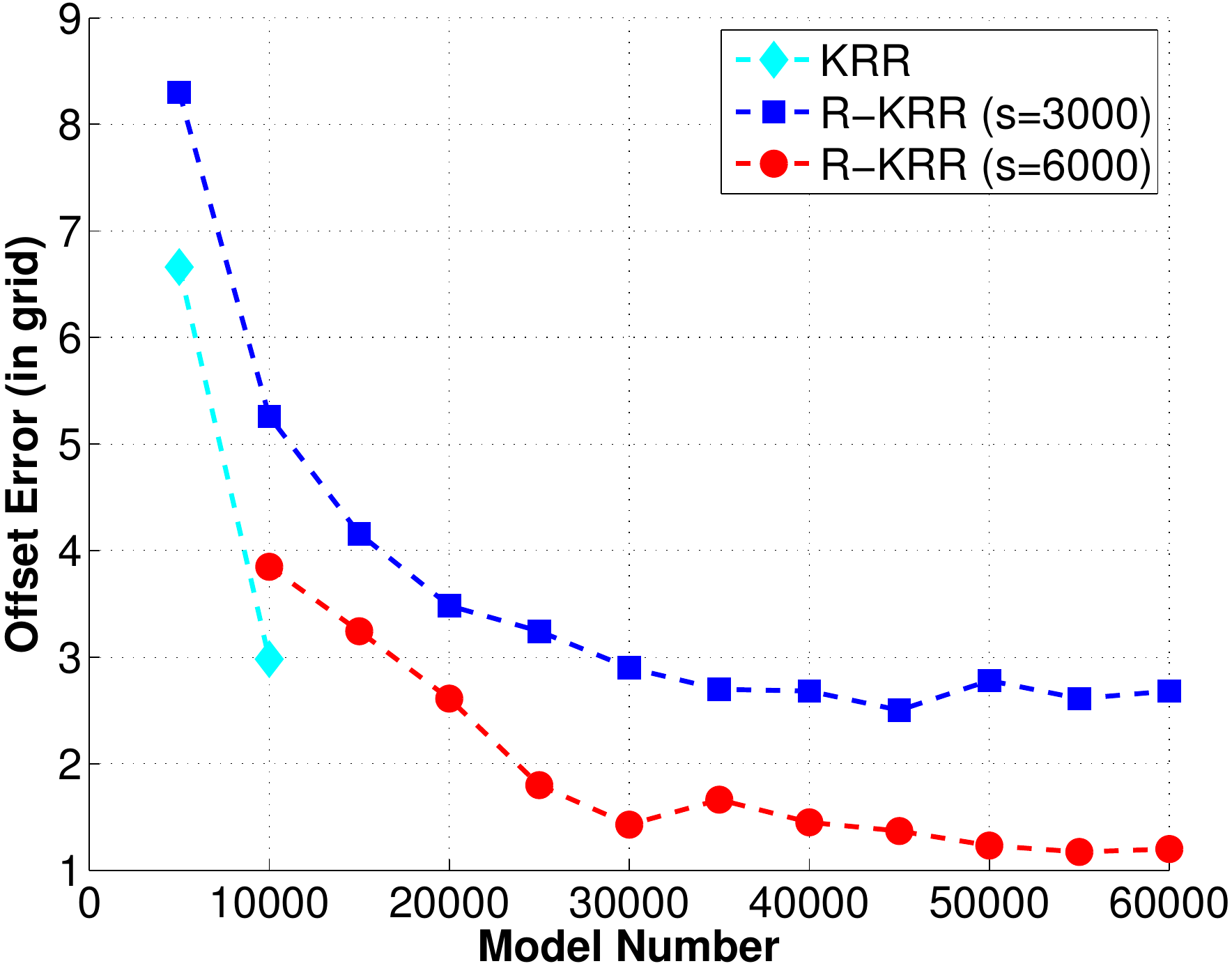}}\vfill
  \caption{Estimation error for (a) dipping angle and (b) offset using conventional KRR method (in cyan), our method 
using $s=3,000$~(in blue), and our method using $s=6,000$~(in red). 
KRR method~(in cyan) yields more accurate results of both angle and offset estimations with small size of data sets, and it fails to provide 
estimation when data sets becomes too large. Our method yields consistently comparable results to KRR on all sizes of data sets.}
\label{fig:perf_accuracy}
\end{figure*}

We provide our first test on the detection accuracy. The estimation result on the dipping angle and horizontal offset 
is provided in Fig.~\ref{fig:perf_accuracy}. We test the performances of our method using two different Nystr\"om 
approximations, $s=3,000$ and $s=6,000$, as well as one other detection approach using conventional KRR method. 
We report the performances of those methods using different sizes of the seismic data. Specifically, we increase the seismic dataset generated from $5,000$ velocity models to $60,000$ velocity models with an incremental of $5,000$ velocity models. 
The corresponding MAE values are reported in Fig.~\ref{fig:perf_accuracy}. In particular, the results of angle estimation is provided in Fig.~\ref{fig:perf_accuracy}(a) and the results of the offset estimation is provided in Fig.~\ref{fig:perf_accuracy}(b). 
We notice when the dataset used for training is small, KRR method~(in cyan) yields more accurate results of both angle and offset estimations. This is reasonable since all the available data sets are used for estimation. After using data from $10,000$ velocity models, KRR method becomes extremely inefficient because of the selection of the parameters using cross-validation. 
It is difficult to evaluate its performance given more training data. While, our method with both Nystr\"om approximations, $s=3,000$ and $s=6,000$, still yields accurate results and efficient performance. In particular, our method with larger Nystr\"om approximation, i.e., $s=6,000$, consistently gives us better results. Our best estimate of the dipping angle on the full seismic data set is $0.5^{\circ}$~(Fig.~\ref{fig:perf_accuracy}(a)). Similarly, we also report the performance of offset estimation in Fig.~\ref{fig:perf_accuracy}(b).  The best estimate of the offset using our method on the full data set is about $1$~grid.

\begin{figure*}
\centering
\includegraphics[width=.50\linewidth]{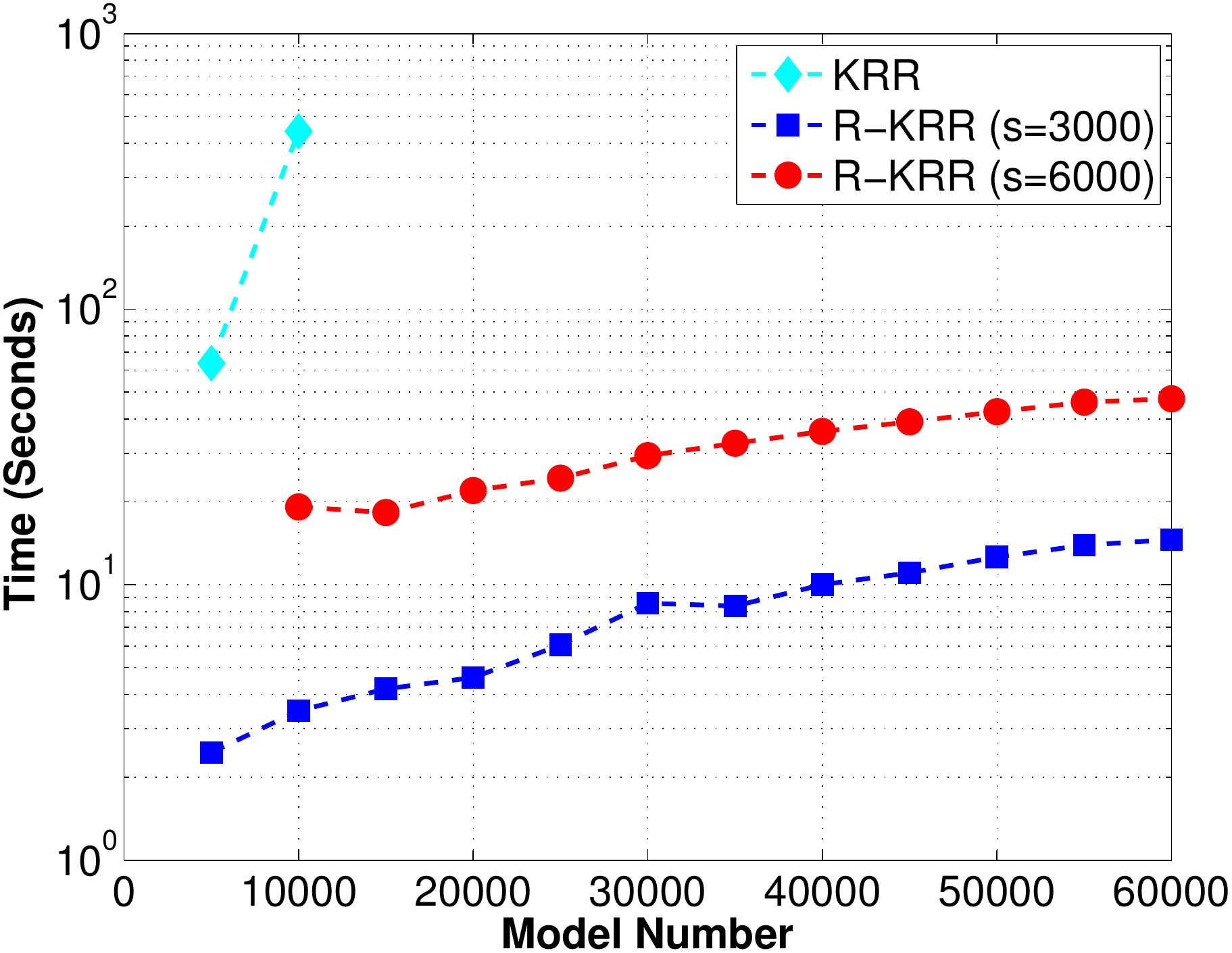}
  \caption{Computational times for the training phase  using our method (in red and blue) and the conventional KRR 
method (in cyan). Our method yields accurate results and is computationally and memory efficient on all data points.}
\label{fig:perf_efficiency}
\end{figure*}

Figures~\ref{fig:perf_efficiency} show the computational time for training phases required by our approach (in blue and red), 
and KRR method (in cyan). Observing Fig.~\ref{fig:perf_efficiency}, KRR approach is much more computationally expensive and memory demanding than our method even if it provides more accurate estimation. On the other hand, our method is significantly more efficient than the conventional KRR method in training when the data sets become large. Utilizing the full dataset, it takes our method on the order of $10$ seconds to train the prediction model. The speed-up ratios between our method and the conventional KRR method in training  phase is up to $1,000$. Such an efficiency would allow the possibility to detect the geologic features in/towards real time. The computational time in estimating the offset is similar to that of the angle estimation and hence we do not report them. Similar conclusions can be drawn that our method not only yields more accurate estimation but also is more computationally efficient out of these two methods.

\begin{figure*}
\centering
\subfloat[]{\includegraphics[width=.45\linewidth]{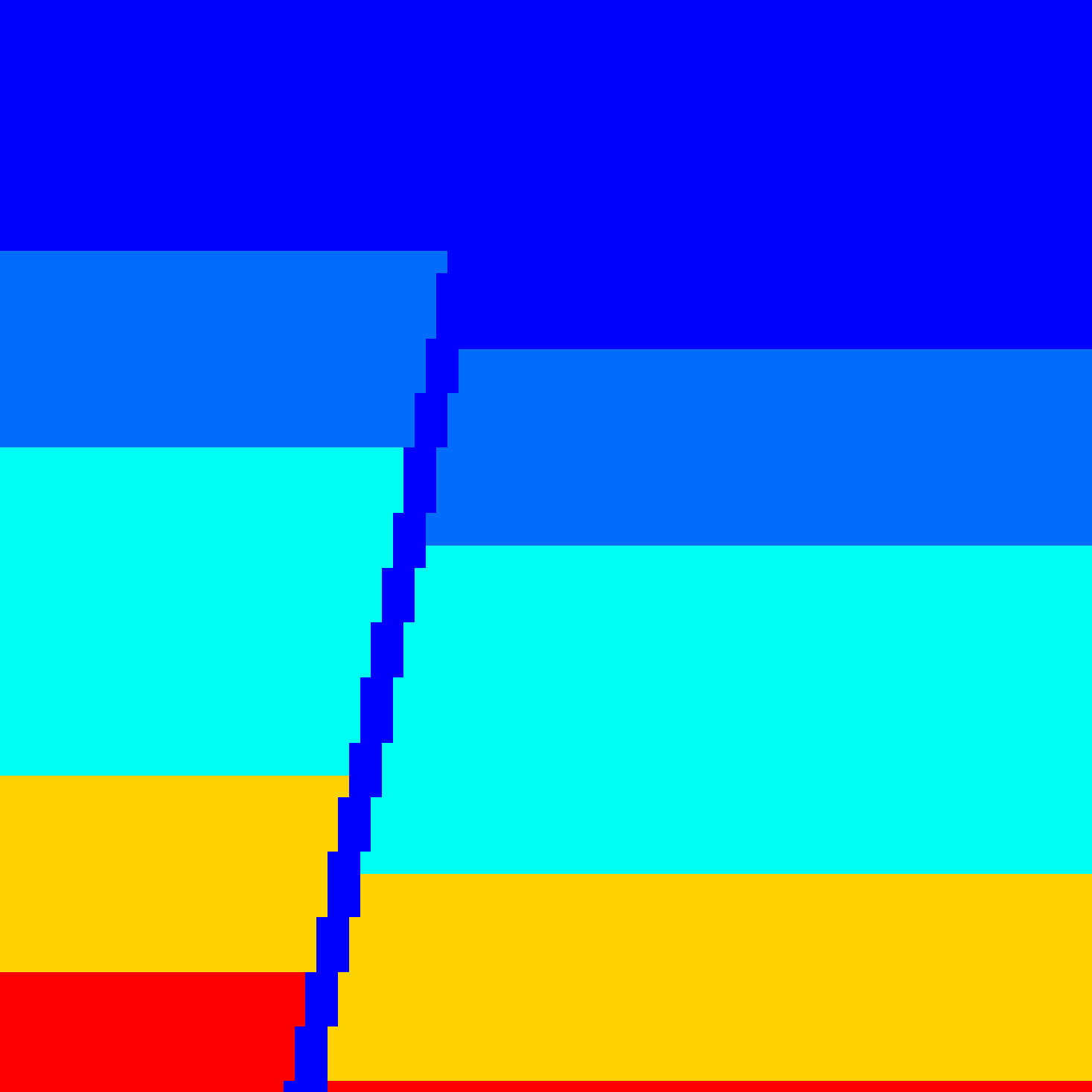}}
\quad \quad \quad \quad
\subfloat[]{\includegraphics[width=.45\linewidth]{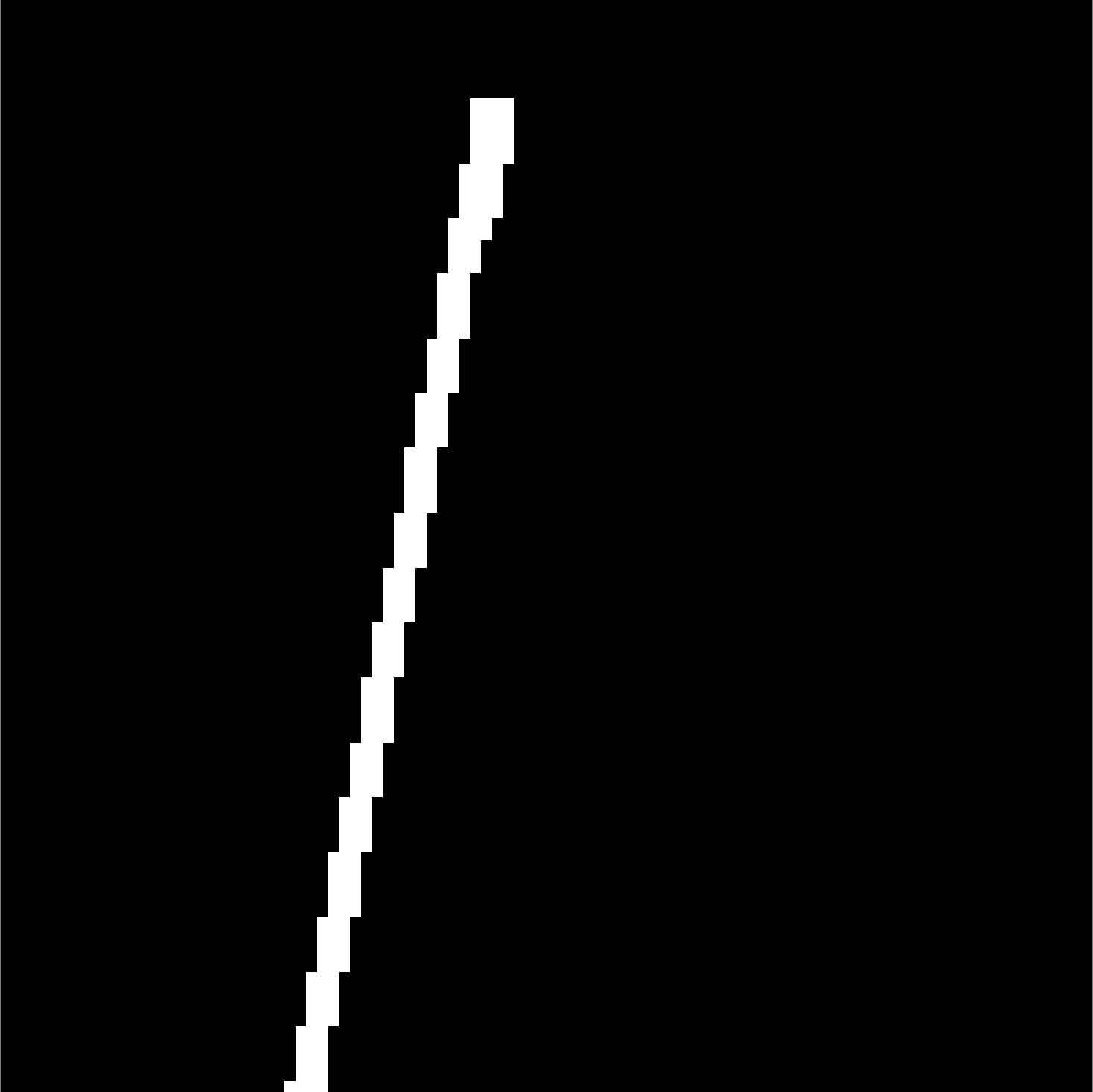}}
\caption{An synthetic model with a geologic fault in it: (a) the true model with angle = $79.1^{\circ}$ and offset = $49.0$; 
        (b) the estimation using our new randomized detection method. The result of our estimation is angle = $79.0^{\circ}$ 
        and offset = $50.0$. Visually, our randomized detection method yields a rather accurate estimation compared to 
        the ground truth. }
\label{fig:EstimationResults}
\end{figure*}

To have a visualization of our estimation, we provide a specific example of the true model and our estimation in Fig.~\ref{fig:EstimationResults}. In Fig.~\ref{fig:EstimationResults}(a), we show our true velocity model  with angle = $79.1^{\circ}$ and offset = $49.0$. The estimation result of our randomized detection method is given in Fig.~\ref{fig:EstimationResults}(b). The result of our estimation is angle = $79.0^{\circ}$ and offset = $50.0$. Visually, our randomized detection method yields a rather accurate estimation compared to the ground truth.

\subsection{Test on the Nystr\"om Sample Size}
\label{sec:TestonRank}

\begin{figure*}
\centering
  \subfloat[]{\includegraphics[width=.48\linewidth]{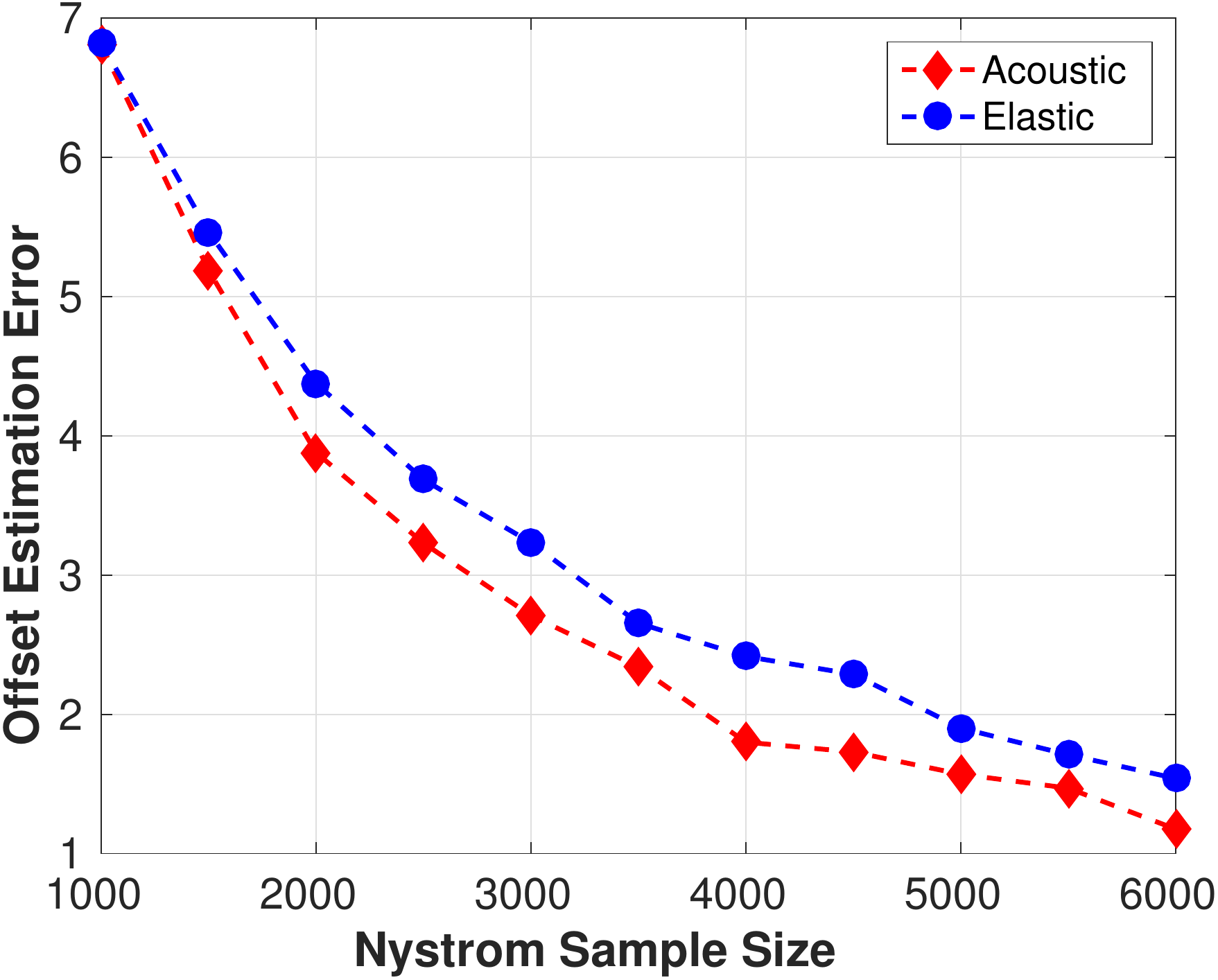}}
   \quad
  \subfloat[]{\includegraphics[width=.48\linewidth]{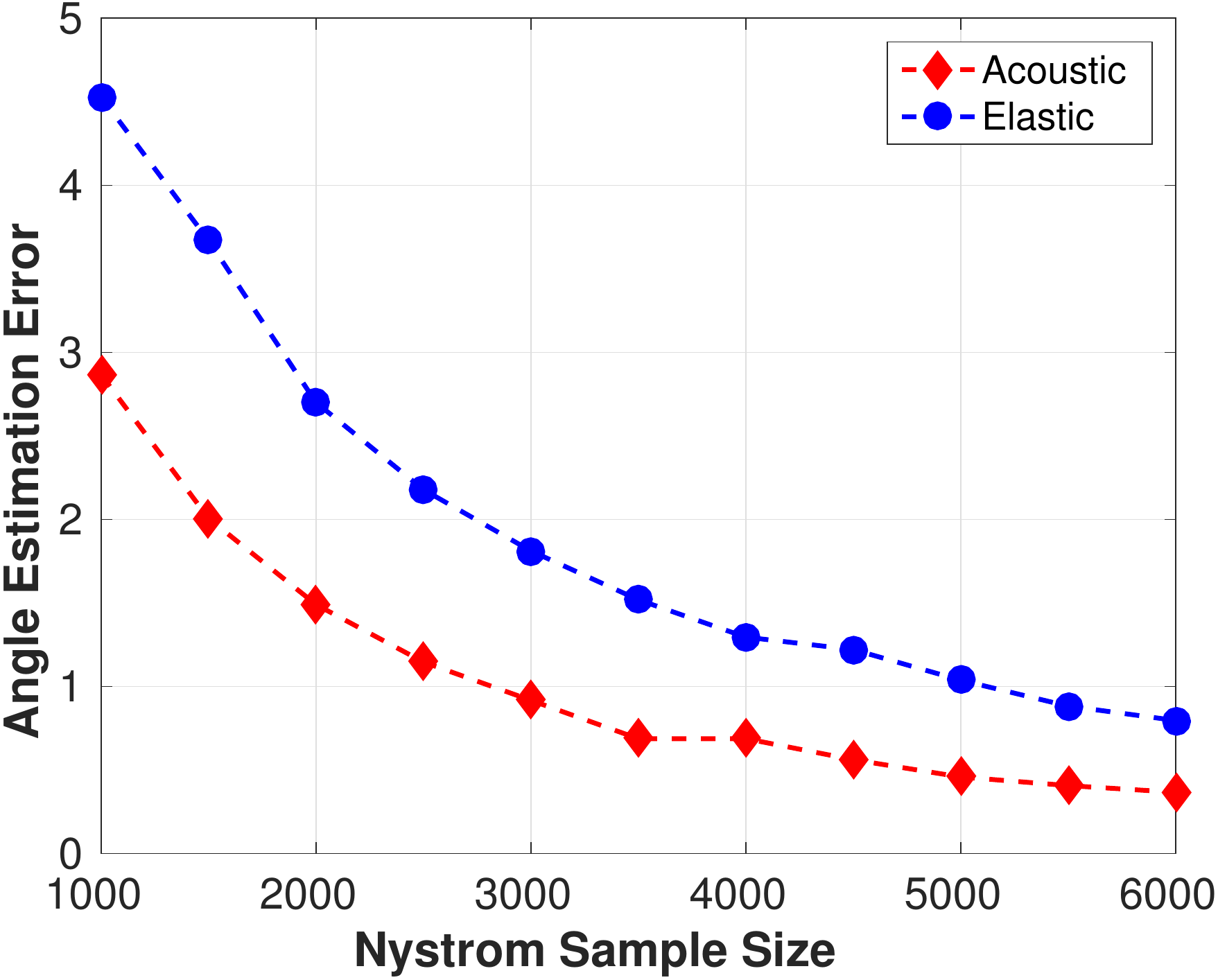}}
  \caption{The estimation results by varying the Nystr\"om sample size, $s$, from from $1,000$ to $6,000$ with an incremental of $500$. Both the estimation results on the horizontal offset~(a) and dipping angle (b) are provided. In both figures, the estimation results using acoustic data sets are plotted in red, and the results using elastic data sets are plotted in blue. We notice that in both figures that with the increase of the Nystr\"om sample size, the estimation accuracy for both dipping angle and horizontal offset also increases, which is due to the fact that more information are utilized in generating the prediction model. }
\label{fig:TestonS}
\end{figure*}

The number of random Nystr\"om sample size, $s$, is critical to the accuracy and efficiency of our randomized feature detection method. The appropriate selection of the Nystr\"om sample size value depends on the redundancy of data sets, which theoretically 
can be justified by the spectrum spanned by the singular vectors of the data sets. In this test, we provide our estimation results by varying the Nystr\"om sample size, $s$, from from $1,000$ to $6,000$ with an incremental of $500$. Besides the acoustic seismic data sets, we also generate elastic seismic data sets for testing our prediction model. The estimation results are provided in Fig.~\ref{fig:TestonS}, where Fig.~\ref{fig:TestonS}(a) is the estimation of horizontal offset and Fig.~\ref{fig:TestonS}(b) is the estimation of the dipping angle. In both figures, the estimation results using acoustic data sets are plotted in red, and the results using elastic data sets are plotted in blue. We notice that in both figures that with the increase of the Nystr\"om sample size, the estimation accuracy for both dipping angle and horizontal offset also increases. This is reasonable and can be explained by the fact that more information are included and utilized for generating the prediction model. Comparing the estimation results using elastic and acoustic seismic data sets, we notice that the one using acoustic seismic data sets yields consistently more accurate results. This is because the elastic models include much more parameters than the acoustic models, which is indicated by Eqs.~\eqref{eq:Forward} and \eqref{eq:ForwardElastic}. With more degree of freedom, more training data sets is therefore needed to achieve the same level of accuracy. To conclude on the selection of the random Nystr\"om sample size, we would suggest to use a value in between $3,000$ to $6, 000$ considering a balance between the accuracy and efficiency.

\subsection{Test on the Randomness of the Nystr\"om Method}
\label{sec:TestonRandomness}

\begin{figure*}
\centering
\includegraphics[width=.60\linewidth]{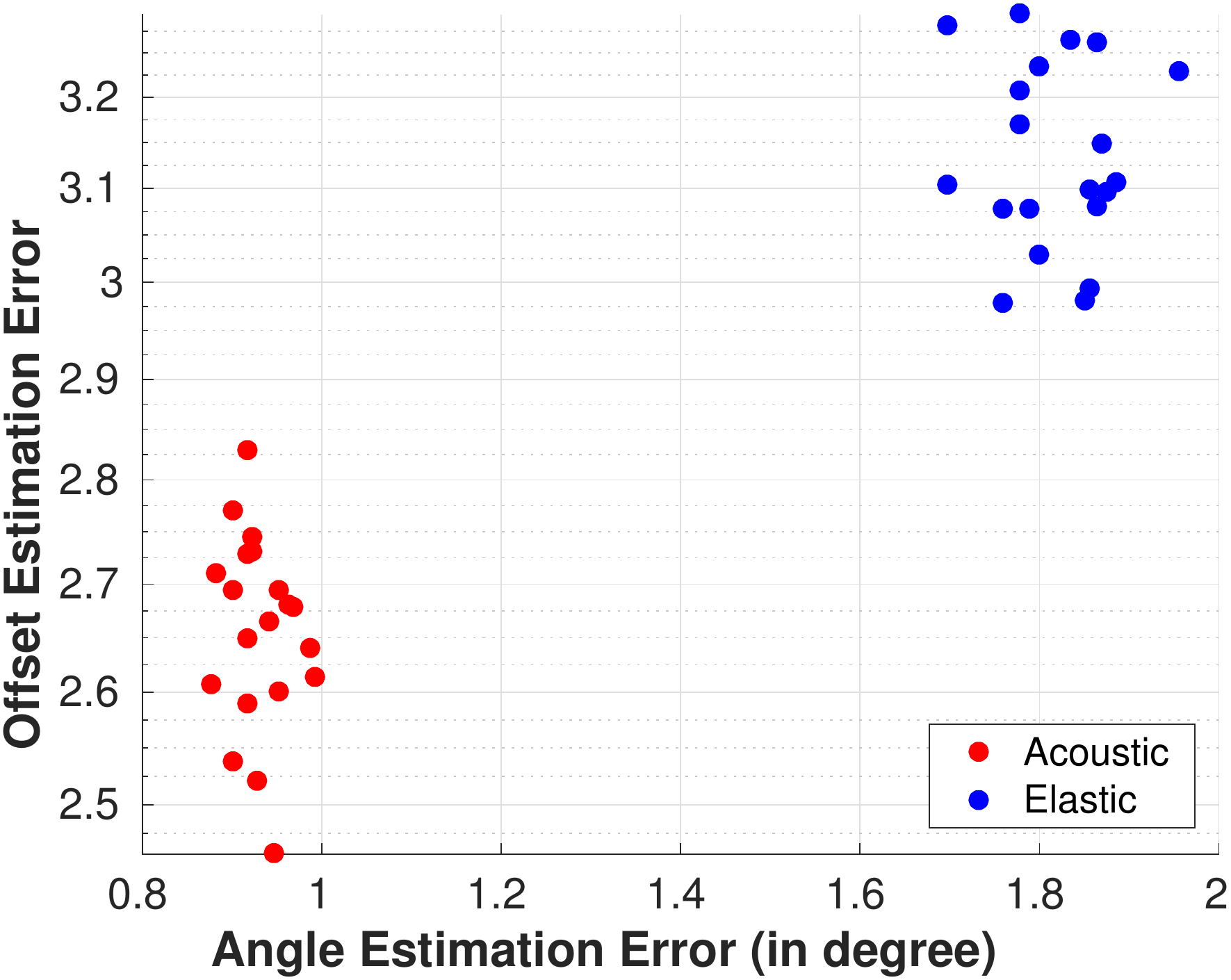}
\caption{ $20$ different realization tests of the Nystr\"om Method are generated. Each of them is drawn from a uniform distribution. Both the dipping angle and horizontal estimation errors according to Eq.~\ref{eq:MAE} are calculated. For all the tests, we set  the Nystr\"om Sample Size, $s=3,000$, and use the full data set size. We report the randomness results in Fig.~\ref{fig:Randomness}, where the acoustic estimation results are plotted in red and the elastic results are plotted in blue. All of the $20$ different realizations lead to similar error estimations of both dipping angle and horizontal offset.}
\label{fig:Randomness}
\end{figure*}

The Nystr\"om method is a randomization-based approach, where the randomness arises from the uniform sampling of the columns in generating the low-rank approximation as in illustrated Fig.~\ref{fig:nystrom}. Here we test the randomness in the prediction made by KRR with Nystr\"om approximation.
We use the same model as in Test~1, where the velocity models is size of $100 \times 100$ grid points. One geological fault zone is contained in the model. One common-shot gather of synthetic seismic data with $32$ receivers is posed at the top surface of the model. We generate $20$ different realization tests of the Nystr\"om method.
Each of them is drawn from a uniform distribution. We calculate their dipping angle and horizontal estimation errors according to Eq.~\eqref{eq:MAE}. For all the tests, we set  the Nystr\"om sample size, $s=3,000$, and use the full data set size. We report the randomness results in Fig.~\ref{fig:Randomness}, where the acoustic estimation results are plotted in red and the elastic results are plotted in blue. We observe that there are two clusters of data points corresponding to the acoustic and elastic scenarios. All of the $20$ different realizations lead to similar error estimations of both dipping angle and horizontal offset. From this test, we conclude that  our method yields robust and accurate results regardless of the randomness nature of the Nystr\"om method.

\section{Conclusions}
\label{sec:Conclusions}

We developed a computationally efficient, machine learning approach for subsurface geological features detection using seismic data sets. 
Instead of detecting geological features from the post-stack image or inversion, our proposed techniques are capable of detecting the geological features of interest from pre-stack seismic data sets. Our data-driven detection methods are based on kernel ridge regression, which can be computationally intensive in training. 
To overcome the issues of excessive memory and computational cost that arises with kernel machines for high dimensional 
dataset, we incorporated a randomized matrix sketching technique. The randomization method can be viewed as a data-reduction 
technique, because it generates a surrogate system that has much lower degrees of freedom than the original problem.
We show through our computational cost analysis that the proposed geologic feature detection method achieves significant reduction 
in computational and memory costs. Furthermore, we utilized a few numerical tests of detecting geological fault zone to demonstrate the performance 
of our detection method. As illustrated by our tests, our method yields comparable  estimation performance compared 
to the  conventional kernel ridge regression. To summarize, our data-driven detection method presents great potential to be utilized for 
detection of subsurface geological features.

\section{Acknowledgments}

This work was supported by the U.S. DOE Fossil Energy research grant. The computation was performed using super-computers 
of LANL's Institutional Computing Program. J. Thiagarajanunder was supported by the U.S. DOE under Contract DE-AC52-07NA27344 to 
Lawrence Livermore National Laboratory.

\appendix
\section{Sherman-Morrison-Woodbury  Matrix Identity}

Given a square invertible $n \times n$ matrix $A$, an $n \times k$ matrix $U$, and a $k \times n$ matrix $V$, let $B$ be an $n \times n$ matrix such that $B = A + UV$. Then, assuming $\left( {{I_k} + V{A^{ - 1}}U} \right) $ is invertible, we have the Sherman-Morrison-Woodbury  matrix identity defined as
\begin{equation}
{B^{ - 1}} = {A^{ - 1}} - {A^{ - 1}}U{\left( {{I_k} + V{A^{ - 1}}U} \right)^{ - 1}}V{A^{ - 1}}.
\label{eq:Woodbury}
\end{equation}

By letting $A=\lambda I_{n}$, $U = \Psi$, and $V = \Psi ^T$ and employing the above formulation to Eq.~\eqref{eq:TildeAlpha1}, we will have 
\begin{eqnarray} 
\hspace{4cm}\tilde{\alpha}
\label{eq:SMW_Deviration1}
& = & \big(\Psi \Psi^T + \lambda I_n \big)^{-1} \mathbf{y},  \\
\nonumber
& = & \left ( (\lambda I_n)^{-1} - (\lambda I_n)^{-1} \Psi (I_k + \Psi ^T (\lambda I_n)^{-1} \Psi) \Psi ^T (\lambda I_n)^{-1} \right )\mathbf{y}, \\
\nonumber& = & \left ( (\lambda ^{-1} I_n - \lambda^{-2} \Psi (I_k + \lambda^{-1} \Psi ^T \Psi) ^{-1} \Psi ^T\right )\mathbf{y}, \\\nonumber
& = & \left ( (\lambda ^{-1} I_n - \lambda^{-2} \lambda \Psi (\lambda I_k +  \Psi ^T \Psi) ^{-1} \Psi ^T\right )\mathbf{y}, \\
& = & \lambda^{-1} y - \lambda^{-1} \Psi (\lambda I_s + \Psi^T \Psi )^{-1} \Psi^T \mathbf{y}
\; \in \; \RB^n.
\label{eq:SMW_Deviration2}
\end{eqnarray}

\bibliographystyle{gji}
\bibliography{reference}

\end{document}